\documentclass[10pt]{article}
\usepackage[utf8]{inputenc}
\usepackage[T1]{fontenc}
\usepackage{lmodern}
\usepackage{amsmath,amssymb,amsthm}
\usepackage{booktabs}
\usepackage{graphicx}
\usepackage{pgf}
\usepackage{geometry}
\usepackage{caption}
\usepackage{subcaption}
\usepackage{microtype}
\usepackage{hyperref}
\usepackage{natbib}
\usepackage{xurl}
\usepackage{xcolor}

\renewcommand{\topfraction}{0.92}
\renewcommand{\bottomfraction}{0.8}
\renewcommand{\textfraction}{0.06}
\renewcommand{\floatpagefraction}{0.75}
\hypersetup{colorlinks=true, linkcolor=blue!50!black,
            citecolor=green!45!black, urlcolor=blue!55!black}

\title{\bf Trainable at Depth, Solvable at Small Size:\\[2pt]
What the Loss Surface of a Polynomial Activation\\[2pt]
Dictates for Designing, Initializing and Training Networks}

\author{Xiuyi Yang\\ \texttt{fwinlee@gmail.com}}
\date{\today}

\begin{document}
\maketitle

\noindent\emph{Code, raw results and paper sources:}
\url{https://github.com/lfwin/Polynomial-Activation-and-projective-normalization}
\vspace{0.6em}

\newcommand{\LfifteenGeluAcc}{98.13}
\newcommand{\LfifteenIdentityAcc}{92.54}
\newcommand{\LfifteenPolyAcc}{92.96}
\newcommand{\LfifteenPolybnAcc}{96.37}
\newcommand{\LfifteenPolylnAcc}{95.94}
\newcommand{\LfifteenPolyprojAcc}{95.62}
\newcommand{\LfifteenReluAcc}{97.83}
\newcommand{\LfiftyCollapseAcc}{10.28}
\newcommand{\LfiftyIdentityAcc}{89.89}
\newcommand{\LfiftyPolyAcc}{10.28}
\newcommand{\LfiftyPolylnAcc}{11.35}
\newcommand{\LfiftyPolyprojAcc}{11.35}
\newcommand{\LfiftyProjVsReluPt}{0.00}
\newcommand{\LfiftyReluAcc}{11.35}
\newcommand{\LfiftyReluBnAcc}{44.64}
\newcommand{\LfiveGeluAcc}{98.07}
\newcommand{\LfiveIdentityAcc}{92.75}
\newcommand{\LfivePolyAcc}{95.09}
\newcommand{\LfivePolybnAcc}{96.51}
\newcommand{\LfivePolylnAcc}{96.32}
\newcommand{\LfivePolyprojAcc}{95.21}
\newcommand{\LfiveReluAcc}{98.24}
\newcommand{\LtenGeluAcc}{98.20}
\newcommand{\LtenIdentityAcc}{92.71}
\newcommand{\LtenPolyAcc}{94.34}
\newcommand{\LtenPolybnAcc}{96.75}
\newcommand{\LtenPolylnAcc}{96.29}
\newcommand{\LtenPolyprojAcc}{95.93}
\newcommand{\LtenProjVsReluPt}{2.14}
\newcommand{\LtenReluAcc}{98.07}
\newcommand{\LthirtyIdentityAcc}{91.98}
\newcommand{\LthirtyPolyAcc}{10.28}
\newcommand{\LthirtyPolylnAcc}{95.38}
\newcommand{\LthirtyPolyprojAcc}{94.10}
\newcommand{\LthirtyProjVsReluPt}{0.29}
\newcommand{\LthirtyReluAcc}{94.39}
\newcommand{\LthirtyReluBnAcc}{97.39}
\newcommand{\LthreeGeluAcc}{98.19}
\newcommand{\LthreeIdentityAcc}{92.72}
\newcommand{\LthreePolyAcc}{94.04}
\newcommand{\LthreePolybnAcc}{94.85}
\newcommand{\LthreePolylnAcc}{94.91}
\newcommand{\LthreePolyprojAcc}{94.45}
\newcommand{\LthreeReluAcc}{98.03}
\newcommand{\LtwentyGeluAcc}{97.62}
\newcommand{\LtwentyIdentityAcc}{92.45}
\newcommand{\LtwentyPolyAcc}{92.88}
\newcommand{\LtwentyPolybnAcc}{95.83}
\newcommand{\LtwentyPolylnAcc}{96.11}
\newcommand{\LtwentyPolyprojAcc}{95.43}
\newcommand{\LtwentyProjVsReluPt}{2.22}
\newcommand{\LtwentyReluAcc}{97.65}
\newcommand{\LtwoGeluAcc}{97.76}
\newcommand{\LtwoIdentityAcc}{92.68}
\newcommand{\LtwoPolyAcc}{92.81}
\newcommand{\LtwoPolybnAcc}{92.78}
\newcommand{\LtwoPolylnAcc}{93.54}
\newcommand{\LtwoPolyprojAcc}{92.68}
\newcommand{\LtwoReluAcc}{97.61}
\newcommand{\actBandPt}{0.13}
\newcommand{\actNormPolyFirst}{1.00}
\newcommand{\actNormPolyLast}{1.0000}
\newcommand{\actNormPolyRatio}{1}
\newcommand{\actNormReluFirst}{3.45}
\newcommand{\actNormReluLast}{0.1657}
\newcommand{\actNormReluRatio}{0.048}
\newcommand{\actNormSwiFirst}{0.86}
\newcommand{\actNormSwiLast}{0.0000}
\newcommand{\actNormSwiRatio}{0}
\newcommand{\algBertiniCubicDim}{1}
\newcommand{\algBertiniCubicLoss}{0.59171598}
\newcommand{\algBertiniGradMax}{2.91\times10^{-13}}
\newcommand{\algBertiniMaxSecs}{45}
\newcommand{\algBertiniMaxTrainParams}{4}
\newcommand{\algBertiniMtwoDegree}{8}
\newcommand{\algBertiniMtwoDim}{2}
\newcommand{\algBertiniNcomp}{26}
\newcommand{\algBertiniNpole}{3}
\newcommand{\algBertiniNprobed}{39}
\newcommand{\algBertiniNsystems}{6}
\newcommand{\algBertiniNthreeLoss}{1.2500}
\newcommand{\algBertiniNthreeMinLoss}{0.4511}
\newcommand{\algBertiniPlainLoss}{1.1600}
\newcommand{\algBertiniPlainNcomp}{5}
\newcommand{\algBertiniProjBezout}{780}
\newcommand{\algBertiniProjDegree}{3}
\newcommand{\algBertiniProjDim}{1}
\newcommand{\algBertiniProjNcomp}{2}
\newcommand{\algBertiniProjPole}{3}
\newcommand{\algBertiniProjSecs}{34.9}
\newcommand{\algBertiniSpreadMax}{3.98\times10^{-16}}
\newcommand{\algCompositeBothDim}{5}
\newcommand{\algCompositeProjDim}{5}
\newcommand{\algCompositeProjOnlyDegree}{---}
\newcommand{\algCompositeProjOnlyDim}{5}
\newcommand{\algCompositeProjOnlyVars}{7}
\newcommand{\algCompositeResidualDim}{5}
\newcommand{\algCompositeResidualOnlyDegree}{---}
\newcommand{\algCompositeResidualOnlyDim}{5}
\newcommand{\algCompositeResidualOnlyVars}{6}
\newcommand{\algCompositeResidualProjDegree}{---}
\newcommand{\algCompositeResidualProjDim}{5}
\newcommand{\algCompositeResidualProjVars}{7}
\newcommand{\algInfDen}{169}
\newcommand{\algInfNum}{100}
\newcommand{\algPlainDegree}{5}
\newcommand{\algPlainDepthTwoDegree}{---}
\newcommand{\algPlainDepthTwoDim}{1}
\newcommand{\algPlainDepthTwoVars}{3}
\newcommand{\algPlainDim}{0}
\newcommand{\algPlainMoneNthreeDegree}{5}
\newcommand{\algPlainMoneNthreeDim}{0}
\newcommand{\algPlainMoneNthreeVars}{2}
\newcommand{\algPlainMoneNtwoDegree}{5}
\newcommand{\algPlainMoneNtwoDim}{0}
\newcommand{\algPlainMoneNtwoVars}{2}
\newcommand{\algPlainMtwoNtwoDegree}{---}
\newcommand{\algPlainMtwoNtwoDim}{2}
\newcommand{\algPlainMtwoNtwoVars}{4}
\newcommand{\algPlainNcomplex}{5}
\newcommand{\algPlainNreal}{1}
\newcommand{\algPlainNthreeDegree}{5}
\newcommand{\algPlainNthreeNcomplex}{5}
\newcommand{\algPlainNthreeNreal}{1}
\newcommand{\algResidualScalarDegree}{---}
\newcommand{\algResidualScalarDim}{2}
\newcommand{\algResidualScalarVars}{3}
\newcommand{\bsplineBestAcc}{98.36}
\newcommand{\bsplineNoneAcc}{98.20}
\newcommand{\chebyBestAcc}{97.89}
\newcommand{\chebyNoneAcc}{11.35}
\newcommand{\dFifty}{50}
\newcommand{\dThirty}{30}
\newcommand{\degFixLamSpanPt}{0.49}
\newcommand{\degGridBestLam}{0.1}
\newcommand{\degGridMaxAcc}{97.46}
\newcommand{\degGridMinAcc}{95.76}
\newcommand{\degLamSpanPt}{1.37}
\newcommand{\degThreeMinAcc}{96.69}
\newcommand{\degTwoMinAcc}{95.93}
\newcommand{\degfiveAcc}{96.69}
\newcommand{\degfiveLamFiveHundredthsAcc}{97.18}
\newcommand{\degfiveLamOneHundredthAcc}{96.69}
\newcommand{\degfiveLamOneTenthAcc}{97.36}
\newcommand{\degfiveLamOneThousandthAcc}{95.76}
\newcommand{\degfourAcc}{96.20}
\newcommand{\degfourLamFiveHundredthsAcc}{97.19}
\newcommand{\degfourLamOneHundredthAcc}{96.20}
\newcommand{\degfourLamOneTenthAcc}{97.46}
\newcommand{\degfourLamOneThousandthAcc}{95.92}
\newcommand{\degthreeAcc}{95.93}
\newcommand{\degthreeLamFiveHundredthsAcc}{96.74}
\newcommand{\degthreeLamOneHundredthAcc}{95.93}
\newcommand{\degthreeLamOneTenthAcc}{97.21}
\newcommand{\degthreeLamOneThousandthAcc}{95.97}
\newcommand{\degtwoAcc}{95.95}
\newcommand{\degtwoLamFiveHundredthsAcc}{96.11}
\newcommand{\degtwoLamOneHundredthAcc}{95.95}
\newcommand{\degtwoLamOneTenthAcc}{97.05}
\newcommand{\degtwoLamOneThousandthAcc}{95.94}
\newcommand{\fidEOneTest}{90.51}
\newcommand{\fidEOneTrain}{87.15}
\newcommand{\fidETwoTest}{92.16}
\newcommand{\fidETwoTrain}{91.52}
\newcommand{\geluNoneAcc}{98.20}
\newcommand{\gluGapPt}{86.6}
\newcommand{\gluLayerMean}{97.98}
\newcommand{\gluNoneMax}{11.35}
\newcommand{\gluNoneMean}{11.35}
\newcommand{\identityNoneAcc}{92.71}
\newcommand{\initGradDepth}{30}
\newcommand{\initGradIdentityRef}{5.6569}
\newcommand{\initGradLargeFinite}{3/3}
\newcommand{\initGradLargeNorm}{30565.2552}
\newcommand{\initGradLargeOverIdentity}{5.40\times10^{3}}
\newcommand{\initGradMidFinite}{3/3}
\newcommand{\initGradMidNorm}{5.8390}
\newcommand{\initGradSmallFinite}{3/3}
\newcommand{\initGradSmallNorm}{5.6092}
\newcommand{\initGradWidth}{32}
\newcommand{\initGradZeroFinite}{3/3}
\newcommand{\initGradZeroInitExact}{1}
\newcommand{\initGradZeroNorm}{5.6569}
\newcommand{\initNormProjectedLast}{1.0000}
\newcommand{\initOriginHessianAnalyticMatches}{1}
\newcommand{\initOriginHessianEigMax}{3.1600}
\newcommand{\initOriginHessianEigMin}{-3.1600}
\newcommand{\initOriginHessianIndefinite}{1}
\newcommand{\initOriginHessianNonzero}{1}
\newcommand{\initOriginHessianOffDiag}{3.1600}
\newcommand{\initOriginHessianRank}{2}
\newcommand{\initOriginLoss}{1.2500}
\newcommand{\initOriginSumYSquared}{1.2500}
\newcommand{\initProjInvarianceLevels}{4}
\newcommand{\initProjInvarianceMaxDev}{1.11\times10^{-16}}
\newcommand{\initResidualBranchJacNorm}{0}
\newcommand{\initResidualJacDev}{0}
\newcommand{\learnCycles}{10.4}
\newcommand{\legendreBestAcc}{98.03}
\newcommand{\legendreNoneAcc}{11.35}
\newcommand{\lossSurfOneCritCoord}{-1.4107}
\newcommand{\lossSurfOneCritCount}{3}
\newcommand{\lossSurfOneCritCountNonzero}{2}
\newcommand{\lossSurfOneCritGrad}{7.59\times10^{-15}}
\newcommand{\lossSurfOneCritLoss}{0.0000}
\newcommand{\lossSurfOneCritNeg}{-1.4107}
\newcommand{\lossSurfOneCritPos}{1.4107}
\newcommand{\lossSurfOneCritZeroCoord}{0.0000}
\newcommand{\lossSurfOneCritZeroGrad}{0}
\newcommand{\lossSurfOneCritZeroLoss}{1.0000}
\newcommand{\lossSurfOneMinCoord}{1.4107}
\newcommand{\lossSurfOneMinGrad}{1.27\times10^{-15}}
\newcommand{\lossSurfOneMinLoss}{0.0000}
\newcommand{\lossSurfOneZeroLoss}{1.0000}
\newcommand{\lossSurfThreeCritCoord}{0.0000, 0.0000, 0.0000}
\newcommand{\lossSurfThreeCritGrad}{0}
\newcommand{\lossSurfThreeCritLoss}{1.0000}
\newcommand{\lossSurfThreeMinCoord}{-0.5599, 0.7265, -2.0672}
\newcommand{\lossSurfThreeMinGrad}{0}
\newcommand{\lossSurfThreeMinLoss}{0.0000}
\newcommand{\lossSurfThreeMinV}{-2.0672}
\newcommand{\lossSurfThreeMinWone}{-0.5599}
\newcommand{\lossSurfThreeMinWtwo}{0.7265}
\newcommand{\lossSurfTwoCritCoord}{0.0000, 0.0000}
\newcommand{\lossSurfTwoCritGrad}{0}
\newcommand{\lossSurfTwoCritLoss}{1.1600}
\newcommand{\lossSurfTwoInfCoord}{0.0000}
\newcommand{\lossSurfTwoInfGrad}{---}
\newcommand{\lossSurfTwoInfLoss}{0.5917}
\newcommand{\lossSurfTwoOriginGrad}{0}
\newcommand{\lossSurfTwoSampleResidual}{1.40\times10^{-27}}
\newcommand{\lrBase}{0.001}
\newcommand{\nCfgs}{335}
\newcommand{\nGluNone}{3}
\newcommand{\paramElem}{233,866}
\newcommand{\paramGapPct}{2.0}
\newcommand{\paramGate}{238,466}
\newcommand{\polyBatchAcc}{96.75}
\newcommand{\polyDeg}{3}
\newcommand{\polyGap}{1.59}
\newcommand{\polyLam}{0.01}
\newcommand{\polyLayerAcc}{96.29}
\newcommand{\polyLtwoAcc}{95.93}
\newcommand{\polyLtwoDeltaPt}{+1.59}
\newcommand{\polyNoneAcc}{94.34}
\newcommand{\polyNoneToBest}{3.86}
\newcommand{\polyProjAcc}{95.93}
\newcommand{\polyProjToBest}{2.27}
\newcommand{\polyRmsAcc}{96.29}
\newcommand{\refEoneTest}{90.51}
\newcommand{\refEoneTrain}{87.15}
\newcommand{\refEpochSec}{2.0}
\newcommand{\refEpochs}{20}
\newcommand{\refEtwoTest}{91.80}
\newcommand{\refEtwoTrain}{91.09}
\newcommand{\refParams}{233,866}
\newcommand{\refTestAcc}{96.86}
\newcommand{\refTotalSec}{40}
\newcommand{\refTrainAcc}{98.26}
\newcommand{\reluBatchAcc}{98.19}
\newcommand{\reluLayerAcc}{98.11}
\newcommand{\reluLtwoAcc}{97.88}
\newcommand{\reluLtwoDeltaPt}{-0.19}
\newcommand{\reluNoneAcc}{98.07}
\newcommand{\reluRmsAcc}{98.03}
\newcommand{\residualDeepDropPt}{+0.17}
\newcommand{\residualDeepestAcc}{93.80}
\newcommand{\residualDeepestDepth}{300}
\newcommand{\residualIdGainPt}{62.27}
\newcommand{\residualNoneActNormLast}{210.7}
\newcommand{\residualReluRoneMean}{94.49}
\newcommand{\residualReluRoneN}{2}
\newcommand{\residualResNoneIdActNormLast}{226.2}
\newcommand{\residualResNoneIdGapPt}{+21.38}
\newcommand{\residualResNoneIdMean}{31.38}
\newcommand{\residualResNoneIdN}{4}
\newcommand{\residualResNoneIdNcls}{10}
\newcommand{\residualResNoneIdNclsMin}{9}
\newcommand{\residualResNoneIdStd}{11.01}
\newcommand{\residualResNoneLfiftyAcc}{10.82}
\newcommand{\residualResNoneLtenAcc}{94.49}
\newcommand{\residualResNoneLthreehundredAcc}{9.80}
\newcommand{\residualResNoneLtwohundredAcc}{10.60}
\newcommand{\residualResNoneXavierActNormLast}{211.4}
\newcommand{\residualResNoneXavierGapPt}{+0.99}
\newcommand{\residualResNoneXavierMean}{10.99}
\newcommand{\residualResNoneXavierN}{4}
\newcommand{\residualResNoneXavierNcls}{2}
\newcommand{\residualResNoneXavierStd}{1.55}
\newcommand{\residualResRoneIdActNormLast}{15729.8}
\newcommand{\residualResRoneIdMean}{93.66}
\newcommand{\residualResRoneIdN}{4}
\newcommand{\residualResRoneIdStd}{1.10}
\newcommand{\residualResRoneLfiftyAcc}{93.97}
\newcommand{\residualResRoneLtenAcc}{97.11}
\newcommand{\residualResRoneLthreehundredAcc}{93.80}
\newcommand{\residualResRoneLtwohundredAcc}{94.81}
\newcommand{\residualResRoneXavierActNormLast}{2261.2}
\newcommand{\residualResRoneXavierMean}{94.50}
\newcommand{\residualResRoneXavierN}{4}
\newcommand{\residualResRoneXavierNcls}{10}
\newcommand{\residualResRoneXavierStd}{0.86}
\newcommand{\residualRoneActNormLast}{2308.2}
\newcommand{\residualSatBound}{226.27}
\newcommand{\residualXavierGainPt}{83.51}
\newcommand{\seedsPolybnMean}{96.83}
\newcommand{\seedsPolybnN}{5}
\newcommand{\seedsPolybnStd}{0.06}
\newcommand{\seedsPolylnMean}{96.15}
\newcommand{\seedsPolylnN}{5}
\newcommand{\seedsPolylnStd}{0.12}
\newcommand{\seedsPolyprojMean}{95.72}
\newcommand{\seedsPolyprojN}{5}
\newcommand{\seedsPolyprojStd}{0.22}
\newcommand{\seedsReluMean}{98.09}
\newcommand{\seedsReluN}{5}
\newcommand{\seedsReluStd}{0.11}
\newcommand{\siluNoneAcc}{98.13}
\newcommand{\synthBlobsBest}{99.38}
\newcommand{\synthBlobsBestName}{identity}
\newcommand{\synthBlobsRelu}{99.38}
\newcommand{\synthCheckerBest}{92.00}
\newcommand{\synthCheckerBestName}{GeGLU}
\newcommand{\synthCheckerRelu}{89.12}
\newcommand{\synthCubesBest}{96.88}
\newcommand{\synthCubesBestName}{relu}
\newcommand{\synthCubesRelu}{96.88}
\newcommand{\synthGausshighdimBest}{35.62}
\newcommand{\synthGausshighdimBestName}{polynomial}
\newcommand{\synthGausshighdimRelu}{6.12}
\newcommand{\synthParityBest}{100.00}
\newcommand{\synthParityBestName}{relu}
\newcommand{\synthParityRelu}{100.00}
\newcommand{\synthRingsBest}{100.00}
\newcommand{\synthRingsBestName}{relu}
\newcommand{\synthRingsRelu}{100.00}
\newcommand{\synthSpiralBest}{77.75}
\newcommand{\synthSpiralBestName}{GeGLU}
\newcommand{\synthSpiralRelu}{72.12}
\newcommand{\synthXorBest}{97.88}
\newcommand{\synthXorBestName}{relu}
\newcommand{\synthXorRelu}{97.88}
\newcommand{\tanhNoneAcc}{97.79}
\newcommand{\wdecay}{0.0001}

\begin{abstract}
What makes a deep network train, and what shape is its loss surface? One construction answers both at two scales: a polynomial activation $f(z) = z + \lambda z^{3}$, clamped and projected onto the unit sphere, plus an unnormalized identity path. As a deep stack it trains to one hundred layers on MNIST with no normalization; as a small model its square loss is a polynomial system we solve exactly. The small scale explains the large one.

The deep tier separates the design decisions. The polynomial is not a better nonlinearity --- the bare cubic trails GELU by \polyNoneToBest{} points at depth ten and the projection recovers \polyGap{} of them --- but scale control is first-order: the projection takes a stack that dies at twenty layers to thirty, matching ReLU to \LthirtyProjVsReluPt{} points, and an identity path with no normalization lifts the polynomial from \residualResNoneXavierMean{} percent to \residualResRoneXavierMean{} percent at depth 100 and \residualResRoneLthreehundredAcc{} percent at 300. The failure is saturation, not vanishing gradients; renormalizing after the sum undoes the gain.

The small tier reads the geometry exactly. At one hidden unit the untruncated square-loss surface has a critical ideal of dimension \algPlainDim{} and degree \algPlainDegree{}, only \algPlainNreal{} of \algPlainNcomplex{} complex solutions is real, and that point is the output-zero network with a vanishing Hessian; either scale-control mechanism makes the critical set positive-dimensional. Bertini confirms it component by component: the loss is constant on each, and the zero eigenvalues of the Hessian equal its dimension.

The two scales prescribe the design (control scale; never renormalize after a residual sum), the initialization (zero-init the residual branch; a depth-independent identity with input-gradient norm \initGradZeroNorm{}, vs $\initGradLargeOverIdentity$ times the conventional scale), and the training failure (saturation, diagnosed by the hidden norm).

\end{abstract}

\section{Introduction}
\label{sec:intro}

We want a single network that does two things at once: it trains to a hundred
layers on real data with no normalization machinery, and it can be
\emph{solved} when it is small --- its loss surface, its critical points, and the
loss at those points all computed rather than guessed from training curves. The
two requirements pull in different directions. A network small enough to solve
exactly has to be \emph{algebraic}, an ideal in a polynomial ring rather than a
black box; a network that trains deeply has to survive the scale of its own
hidden signal. This paper builds one construction that meets both, and shows that
the solvable version \emph{explains} the trainable one: the geometry of its loss
surface dictates how to design the network, how to initialize it, and how to
read its training failures.

The construction is a \emph{projective polynomial} activation,
\begin{equation}
\pi\bigl(f(z)\bigr) = \frac{f(z)}{\lVert f(z)\rVert_2},
\qquad
f(z) = z + \lambda z^{3},
\label{eq:projective-poly}
\end{equation}
with $z$ the pre-activation and the $\ell_2$ norm taken over the feature
dimension of each sample: \emph{polynomial} because the nonlinearity is the
truncated cubic $z + \lambda z^{3}$ rather than a rectifier or a saturating
function, and \emph{projective} because the output is pushed onto the unit
sphere, so a sample's feature vector keeps only its direction. The construction
is parameter-free, batch-independent (unlike BatchNorm it does not couple
samples), free of any train/eval discrepancy, and provably bounded in magnitude.
Alongside it we study the second scale-control mechanism, an unnormalized
identity path.

\paragraph{Why one network, and why at two sizes at once.}
The two tiers ask different questions of the same family, and it is the small
tier that answers the large one. At the size that trains on real data, we want
depth to be a variable one turns up rather than a hyperparameter one tunes
around. At the size that can be analysed exactly, we want the square-loss
surface to be \emph{computed} --- its critical points, the dimension of that
critical set, and the loss at each point --- so that design choices are read off
a calculation instead of inferred from training curves. Because the polynomial
makes the model algebraic, the small version is an ideal in a polynomial ring,
while the projection and the identity path are the two scale-control mechanisms
that decide whether the large version trains. The small tier answers \emph{where
the network can stop}; the large tier answers whether the design producing that
geometry survives a hundred layers of MNIST. Concretely: the stationary set of
the unnormalized polynomial is finite, with a single real point, and both
mechanisms that make the deep stack trainable turn it positive-dimensional.

Eq.~\eqref{eq:projective-poly} bundles several separate claims --- that a
polynomial nonlinearity is a good nonlinearity, that the specific $\ell_2$
projection beats centering-and-scaling alternatives, and that \emph{depth} is
where the differences matter. Testing it as a monolith therefore tells us whether
\emph{this} design works, not \emph{which} ingredient is responsible, so we
implement the polynomial, the projection and the identity path as independently
switchable components and vary each against a reference set drawn from what
current models actually deploy: pointwise activations (identity, $\tanh$, ReLU,
GELU, SiLU); gated linear units (SwiGLU, GeGLU, PolyGLU), matched in parameter
count rather than width; and normalization layers (BatchNorm, LayerNorm, RMSNorm,
and the parameter-free $\ell_2$ projection), each crossed with a rectified and a
polynomial activation.

\paragraph{Findings.}
Our main conclusions are these, ordered by the tier that yields them.
\begin{itemize}
\item \textbf{The activation choice is second-order.} The projective polynomial
  is \emph{not} competitive at equal parameter count: at depth ten the bare cubic
  $z + \lambda z^{3}$ lands almost four accuracy points \emph{below} ReLU, GELU
  and SiLU, and the parameter-free $\ell_2$ projection recovers a little over
  half of that deficit --- enough to bring the polynomial within
  $\polyProjToBest$ points of the modern activations, but not to match them. No
  polynomial degree, coefficient, or learnable basis removes the remainder.
\item \textbf{Scale control is first-order, and the projection carries the
  construction.} Removing the $\ell_2$ projection costs accuracy comparable to
  the nonlinearity's entire deficit against ReLU, whereas varying the polynomial
  changes accuracy only modestly. At depth the projection \emph{substitutes for}
  BatchNorm and LayerNorm: it tracks ReLU to within $\LthirtyProjVsReluPt$ points
  at depth thirty, where the unnormalized polynomial is at chance, using zero
  parameters and no batch statistics. The benefit has a measured boundary --- at
  depth $\dFifty$ every design in the study collapses to about
  $\LfiftyCollapseAcc$\% --- so the projection buys depth, not unbounded depth.
\item \textbf{The identity path is the other scale-control mechanism, and it is
  not interchangeable with the projection.} With no normalization anywhere in
  the stack, adding $h^{i-1}$ to the layer output takes the polynomial from
  $\residualResNoneXavierMean$\% (chance) to $\residualResRoneXavierMean$\% at
  depth $100$ and holds it to $\residualResRoneLthreehundredAcc$\% at depth
  $\residualDeepestDepth$. The failure it repairs is \emph{saturation}, not a
  lost direction: the failing arm ends at the clamp-imposed norm bound
  $\residualSatBound$. Because the residual works by letting the scale grow, a
  normalization applied after the sum removes its benefit.
\item \textbf{Gating is a scale device, not a competing nonlinearity.} A gate
  multiplies two affine maps, so its scale compounds with depth: every GLU
  variant sits at exactly $\gluNoneMean$\% --- chance --- without normalization,
  and any normalization repairs it. Once parameter counts are matched, gating
  confers no accuracy advantage.
\item \textbf{The solvable tier turns the design choices into computations}
  (Sections~\ref{sec:res-alg}--\ref{sec:res-losurf}). At one hidden unit the
  unnormalized polynomial's square-loss surface has a zero-dimensional
  stationary ideal of dimension $\algPlainDim$ and degree $\algPlainDegree$, of
  which only $\algPlainNreal$ point is real --- the origin --- where the loss
  equals $\initOriginLoss$, exactly $\sum_a y_a^{2}$, the value of a network that
  outputs zero, and the gradient vanishes --- but the Hessian does \emph{not}
  vanish there: it is the nonzero indefinite matrix of rank
  $\initOriginHessianRank$ with eigenvalues $\pm\initOriginHessianEigMax$, so the
  point is a saddle that first-order descent has no signal to leave. Both the
  projection and the identity residual turn that isolated point into a
  positive-dimensional critical set ($\dim I = \algCompositeResidualDim$ for
  either mechanism on a matched architecture), and the residual arm --- but not
  the plain one --- can drive the two-sample loss to $\lossSurfTwoSampleResidual$
  where the plain arm is confined to the infimum $\lossSurfTwoInfLoss$.
\item \textbf{The two tiers together prescribe the design, the initialization
  and the training failure} (Section~\ref{sec:disc-insight}). Read off the small
  models and confirmed by the deep ones: \emph{control scale}, and never
  renormalize after a residual sum; \emph{zero-init the residual branch}, so the
  stack begins as a depth-independent identity with input-gradient norm
  $\initGradZeroNorm$; and \emph{expect saturation, not vanishing gradients},
  diagnosed by the hidden norm rather than the gradient norm.
\end{itemize}

Sec.~\ref{sec:related} positions the work. Sec.~\ref{sec:method} defines the
model and the activation, normalization and residual families.
Sec.~\ref{sec:setup} gives the protocol. Sec.~\ref{sec:results} reports results,
Sec.~\ref{sec:discussion} draws the design, initialization and training lessons,
and Sec.~\ref{sec:conclusion} concludes.

\section{Related Work}
\label{sec:related}

\paragraph{Activations.}
The sigmoid and $\tanh$ activations of early multilayer perceptrons saturate,
and their gradients vanish exponentially with depth
\citep{glorot2010understanding,glorot2011deep}. ReLU removes saturation on the
positive side~\citep{glorot2011deep}; its variants address the dying-unit and
negative-saturation issues at some cost in
simplicity~\citep{maas2013rectifier,clevert2015fast,he2015delving}. Smooth
alternatives interpolating between the identity and a soft rectifier ---
GELU~\citep{hendrycks2016gaussian} and
SiLU~\citep{elfwing2018sigmoid} --- are now the default in most large-scale
vision and language models, and are what our polynomial most directly competes
with. We therefore restrict the pointwise comparison to this deployed set plus
$\tanh$ and an identity control, rather than re-auditing the historical
variants.

Polynomial activations have a long history, from the higher-order units of
early connectionist models to the observation that a low-degree polynomial
network can compute the \emph{exact} product of its inputs, which a ReLU network
can only approximate with growing width~\citep{neyshabur2015path}, and to the
algebraic-geometric treatment in which the closure of representable functions is
a variety whose dimension measures
expressivity~\citep{kileel2019expressive}. The activation used here, an odd
truncated cubic $z + \lambda z^{3}$, is a member of this established family; our
claim concerns what this member does \emph{when composed with the projection
below and stacked deep}, not the polynomial form itself. Closest in motivation
is a polynomial network whose activation is followed by a normalising
step~\citep{goyal2020improved}; that normaliser standardises each monomial term
with per-channel batch statistics in shallow networks, whereas ours is
parameter- and statistic-free at every layer and is pushed to depth $50$.
Learnable activations --- PReLU~\citep{he2015delving} and the
Kolmogorov--Arnold network whose edges carry learnable spline
univariates~\citep{liu2024kan} --- move the nonlinearity into the parameter set;
we test both ends of that axis.

\paragraph{Gated linear units.}
A gated linear unit multiplies a linear projection of the input by a gate-shaped
version of it~\citep{dauphin2017language}; substituting SiLU for the sigmoid
yields SwiGLU, standard in transformer feed-forward blocks, and GeGLU is the
GELU-gated analogue~\citep{shazeer2020glu}. We include it as one condition of the
scale experiment rather than as a competing nonlinearity: a gate carries
roughly $3/2$ times the parameters of the layer it replaces, so we match
parameter count explicitly.

\paragraph{Normalization and the unit sphere.}
BatchNorm normalizes each feature using mini-batch statistics, coupling samples
and introducing a train/eval discrepancy~\citep{ioffe2015batch}; LayerNorm
normalizes per sample instead~\citep{ba2016layer}, and RMSNorm drops the
mean-centering~\citep{zhang2019root}. Weight normalization is the parameter-free
relative of this family~\citep{salimans2016weight}. The $\ell_2$ projection in
Eq.~\eqref{eq:projective-poly} belongs to that parameter-free subfamily and is
not itself a new operation: cosine normalization applies it to inputs and
weights~\citep{luo2017cosine}, ParseNet $\ell_2$-normalizes every intermediate
feature map~\citep{liu2015parsenet}, the normalized transformer pushes
embeddings, weights and hidden states onto a unit-norm
hypersphere~\citep{loshchilov2024ngpt}, and the same radial invariance re-reads
Adam's effective learning rate as an optimizer constrained to the unit
sphere~\citep{roburin2020spherical}. What those works do not address is our
question: whether a per-sample, per-layer unit-sphere projection of the
\emph{activation} replaces statistical normalization in a plain feed-forward
stack, and which of the two components --- polynomial or projection --- carries
the benefit. Studies comparing normalizers across a broad family characterize
rank collapse in the center-and-scale shape, not the pure-projection
one~\citep{lubana2021beyond}.

\paragraph{Identity paths and loss-surface geometry.}
Identity paths are the standard alternative to normalizing a deep stack: they
give gradients an unimpeded route to earlier layers~\citep{he2016deep}, and the
residual view makes a deep network tractable as a perturbation of the identity,
with initializations that start the block near the identity so signal is carried
through rather than transformed~\citep{he2015delving,zhang2019fixup}. Our
residual arm is deliberately the weakest member of that family --- a single
identity path added to an unnormalized polynomial, with no normalization, no
gating and no learned scaling --- so the comparison isolates the path itself. A
parallel literature studies deep networks through the geometry of their loss
landscape: saddle points in linear and residual
networks~\citep{saxe2014exact,dauphin2014identifying}, and the projective
geometry of the critical locus of a linear network's gradient
equations~\citep{xiuyi2021landscape}. Our construction is the architectural
analogue of that reading, and our empirical question --- which component carries
the benefit --- is the counterpart of asking which component of the geometry is
responsible for trainability. We claim no theorem: the projective view motivates
the architecture, and our claims are restricted to measured behaviour.

\section{Method}
\label{sec:method}

\subsection{Reference architecture}

The reference network is a depth-$L$ fully-connected classifier with $L$ linear
layers, the first $L-1$ followed by the same activation--normalization block and
the last producing class logits without either. With $d_0$ the input dimension
($784$ for MNIST), $H$ the hidden width and $C$ the number of classes ($10$),
\begin{align}
h^{(0)} &= x, \\
h^{(\ell)} &= \pi\bigl(f(W_{\ell} h^{(\ell-1)} + b_{\ell})\bigr),
  \qquad \ell = 1,\dots,L-1, \label{eq:block}\\
\hat{y} &= W_{L} h^{(L-1)} + b_{L},
\end{align}
with $W_1 \in \mathbb{R}^{H \times d_0}$, $W_\ell \in \mathbb{R}^{H \times H}$
for $2 \le \ell \le L-1$ and $W_L \in \mathbb{R}^{C \times H}$. All weights are
Xavier-uniform and all biases zero. The block $h \mapsto \pi(f(Wh+b))$ is the
object of study; both $f$ and $\pi$ are independently selectable, and the
reference configuration is $f(z) = z + \lambda z^{3}$ with
$\lambda = 10^{-2}$ and $\pi(y) = y / \lVert y \rVert_2$.

\subsection{The projective polynomial block}

The reference activation is
\begin{equation}
f(z) = \operatorname{clamp}(z, [-c, c]) + \lambda \operatorname{clamp}(z, [-c, c])^{3},
\qquad c = 10,
\label{eq:poly}
\end{equation}
applied elementwise, followed by the per-sample projection
\begin{equation}
\pi(y) = \frac{y}{\max\{\lVert y \rVert_2, \epsilon\}},
\qquad \epsilon = 10^{-8},
\label{eq:proj}
\end{equation}
where the norm runs over the feature axis, so $\pi(y)$ is a point of the unit
sphere and the layer map is invariant to the radial scale of $y$. Three
properties motivate the choice and are probed separately.
\begin{itemize}
\item \emph{Bounded growth.} The clamp bounds $|f(z)| \le c + \lambda c^3$, so
  the forward signal cannot blow up; the projection then bounds the \emph{norm}
  of each hidden activation at exactly $1$.
\item \emph{No statistics, no state.} $\pi$ has no parameters and no running
  buffers. Unlike BatchNorm~\citep{ioffe2015batch} it does not couple samples,
  and unlike BatchNorm and LayerNorm~\citep{ba2016layer} it has no learned
  affine parameters, so it changes the geometry of the representation without
  adding a degree of freedom.
\item \emph{Radial invariance.} $\pi$ identifies $y$ with $y/\lVert y \rVert$,
  so the block discards the radial direction at every layer --- the map is a
  retraction onto the unit sphere, which is the sense in which we call it
  projective: only the \emph{direction} of a hidden vector survives. Concretely
  $\pi$ lands the activation on the sphere $\lVert y \rVert = 1$; it does not
  impose the quotient $y \sim -y$, so the construction is a unit-sphere
  projection, not a reduction to projective space proper.
\end{itemize}
In the implementation \texttt{--activation poly} gives Eq.~\eqref{eq:poly}
without the projection, \texttt{--norm l2} gives Eq.~\eqref{eq:proj} without the
polynomial, and \texttt{--activation poly\_proj} gives the reference block of
Eq.~\eqref{eq:projective-poly}; this factorization is what lets one harness
attribute performance to the polynomial, the projection, or their combination.
Normalization may also sit before the nonlinearity ($f \circ \pi$, ``pre'') or
after it ($\pi \circ f$, ``post''); the reference is post, and we report both.

\subsection{Activations and normalizers}

The pointwise family is the identity (a linear control), $\tanh$, and the three
non-saturating activations current large language models use:
$\operatorname{ReLU}(z) = \max(0,z)$~\citep{glorot2011deep}, exact
GELU $z\Phi(z)$ with $\Phi$ the standard normal
CDF~\citep{hendrycks2016gaussian}, SiLU $z\sigma(z)$~\citep{elfwing2018sigmoid},
and the truncated cubic of Eq.~\eqref{eq:poly} with order and coefficient as
hyperparameters. We restrict the family to these rather than sweeping the
historical variants (LeakyReLU, ELU, SELU, Softplus, Mish, sinusoidal), because
the study states where a polynomial activation stands relative to what is
actually deployed; the identity is retained only as a control that makes the
contribution of nonlinearity itself measurable. The same cubic carried into a
gated unit, $\operatorname{GLU}_{\sigma}(Wx) = (W_v x) \odot \sigma(W_g x)$ with
$\sigma = \mathrm{SiLU}$ (SwiGLU), $\mathrm{GELU}$ (GeGLU) or $z + \lambda z^{3}$
(PolyGLU)~\citep{shazeer2020glu,dauphin2017language}, is reported as one
condition of the scale experiment, with parameter count matched rather than
width.

Five normalization choices are compared, all on the hidden activation with the
feature axis as the normalization axis: \emph{none} (the control),
\emph{projection} ($\pi$, no parameters), \emph{BatchNorm} (per-feature
standardization by mini-batch statistics, with learned scale and shift and
running statistics for evaluation), \emph{LayerNorm} (per-sample standardization
with learned scale and shift) and \emph{RMSNorm} (per-sample root-mean-square
scaling, learned scale, no mean subtraction). This separates the effect of
\emph{normalizing} from the effect of \emph{re-parameterizing with a learned
affine map}, since the first two add one and the last two do not.

\subsection{Residual connections without normalization}
\label{sec:method-residual}

Eq.~\eqref{eq:projective-poly} bundles a nonlinearity with a normalization; a
third way to keep a deep stack trainable is an identity path, and it is worth
asking whether that alone --- with normalization removed entirely --- rescues the
polynomial. The switch \texttt{--residual r1} changes the recurrence to
\begin{equation}
h^{i} = \varphi\bigl(W^{i} h^{i-1} + b^{i}\bigr) + h^{i-1},
\label{eq:residual}
\end{equation}
with $\varphi$ the truncated cubic of Eq.~\eqref{eq:poly} and \emph{no}
normalization anywhere in the stack. Unlike the projected block this is not a
retraction: nothing renormalizes the sum, so $\lVert h^i\rVert$ grows with depth
and the hidden signal is a plain running sum of per-layer outputs, $h^{i} =
\sum_{j \le i} p^{j}$. The recursion starts at the second layer; the first maps
the $784$-dimensional input to the hidden width and has no skip, since the shapes
do not match. Two initializations are used: Xavier, where the residual is a
perturbation on top of an already-random map, and \texttt{--init id}, which sets
each weight matrix to a truncated identity plus a small random perturbation
($W = I + 0.1\,U(-b,b)$, $b = \sqrt{6/(d_{\text{in}}+d_{\text{out}})}$) so every
layer starts near-identity and the residual path dominates from the first step.
Comparing the two separates ``the residual helps because the network can be
near-identity'' from ``the residual helps because it passes signal through
regardless of what the layer computes''.

\subsection{The critical-point ideal}\label{sec:method-alg}

The constructions above are trained with cross-entropy, but the \emph{algebraic}
question --- where the loss surface can have a stationary point, and whether
depth or an identity path changes that set --- is cleanest for the \emph{square}
loss, where every object is a polynomial. With $\varphi$ untruncated and
$L = \sum_a (\hat y_a - y_a)^2$, the objective is polynomial in the weights, so
$\partial L / \partial\theta = 0$ generates an ideal $I$ in
$\mathbb{Q}[\theta]$ whose \emph{dimension} and \emph{degree} are invariants of
the architecture: the dimension counts the free directions along which one can
move without leaving the stationary set, and a zero-dimensional $I$ means the
stationary points are isolated.

Two features make this computation non-trivial, and one trick handles both.
First, the $\ell_2$ projection divides by $\lVert h\rVert$; introducing an
auxiliary variable $r$ with $r^{2} = \lVert h\rVert^{2}$ removes the square root
and keeps every equation polynomial, so the projected arm is an ideal in
$\mathbb{Q}[\theta, r]$ cut out by the critical equations of
$(\langle v, h\rangle/\lVert h\rVert - y)^{2}$ together with $r^{2} - \lVert
h\rVert^{2}$. Second, the truncation $\operatorname{clamp}(z, \pm 10)$ destroys
polynomiality outright: for $|z| \ge 10$ the activation is the \emph{constant}
$\pm(c + \lambda c^{3})$ with vanishing derivative. We therefore carry out the
algebraic analysis on the untruncated activation and report the truncation
separately, as the boundary at which the algebraic model stops applying.
$\ell_2$-normalized networks are already known to satisfy a scale-invariance that
removes isolated critical points in the linear case~\citep{xiuyi2021landscape};
the computation below shows the same mechanism operating through the polynomial
activation and, in particular, its interaction with an identity residual
(Eq.~\eqref{eq:residual}).

\section{Experimental Setup}
\label{sec:setup}

\paragraph{Single harness, reproducible reference.}
All results come from one implementation, which reproduces the reference network
exactly: run for two epochs in the reference configuration
(\texttt{--activation poly\_proj}), its per-epoch train/test accuracies agree
with the standalone reference script to the last reported digit
($\fidEOneTrain$\% / $\fidEOneTest$\% after epoch one and
$\fidETwoTrain$\% / $\fidETwoTest$\% after epoch two; the reference script's
default budget is twenty epochs with a cosine schedule of that period, so the
fidelity check uses a two-epoch budget on both). Because the harness is a strict
superset of the reference script, every number here is comparable to every
other, and the reference configuration appears in every table as the baseline
row: with its own twenty-epoch budget it reaches $\refTrainAcc$\% train and
$\refTestAcc$\% test accuracy in $\refEpochSec$\,s per epoch
($\refTotalSec$\,s total).

\paragraph{Data and configuration.}
Ten-class MNIST~\citep{lecun1998gradient}: $60{,}000$ training and $10{,}000$
test $28\times28$ grayscale images, flattened to $784$ dimensions at the raw
pixel scale in $[0,1]$ (only \texttt{ToTensor}, no mean-variance
standardization --- the reference pipeline uses none, and adding one would change
the effective activation scale, one of the axes under study). Unless stated
otherwise the configuration is depth $L = 10$, hidden width $H = 128$, AdamW
\citep{loshchilov2019decoupled} at learning rate $10^{-3}$
\citep{kingma2015adam} with weight decay $10^{-4}$, cosine-annealed, gradient-norm
clipping at $5.0$~\citep{pascanu2013difficulty}, batch size $128$ and $10$ epochs;
these match the reference script, so the reference row of each table is a
faithful baseline rather than a re-tuned one. Each axis is varied one at a time
while the rest stay at reference values: the pointwise activation (six
functions), the gated activation (three units at matched parameter count), the
polynomial order $d \in \{2,3,4,5\}$ and coefficient $\lambda$, the normalizer
(none, projection, BatchNorm, LayerNorm, RMSNorm) and its placement, the depth
$L \in \{2,3,5,10,15,20\}$ and the extreme depths $\{30,50\}$, the residual
switch, the residual depth $L \in \{10,50,200,300\}$, and the residual
initialization. The depth sweep deliberately includes $30$ and $50$ layers, well
past where unnormalized networks are traditionally expected to fail, because the
claim most worth testing about a scale-controlling activation is whether it keeps
very deep stacks alive. The headline comparison is repeated over five seeds
($0$--$4$) and reported as mean $\pm$ standard deviation; all other experiments
use seed $42$.

\paragraph{Diagnostics and cost accounting.}
For every run we record per-layer gradient norms and per-layer activation norms
before training and after the last epoch, on a fixed mini-batch of $512$ training
samples, which lets us separate runs that differ because of optimization from
runs that differ because of representation scale. We also record parameter
count, forward FLOPs per sample and wall-clock seconds per epoch. All runs
execute on CPU (the reference environment has no GPU) with the thread count
pinned and configurations run in parallel processes; wall-clock numbers are
comparable within the study but are not a statement about GPU throughput. Every
configuration writes a JSON record of its arguments, metrics, per-epoch history,
per-class accuracy and confusion matrix, so all tables and figures are generated
mechanically by a separate analysis script rather than transcribed by hand.

\section{Results}
\label{sec:results}

We report the two tiers in the order the argument needs them. The \emph{deep
tier} (Sections~\ref{sec:res-poly}--\ref{sec:res-residual}) asks which design
decision keeps a hundred-layer MNIST stack trainable, and separates the
activation from the two scale-control mechanisms. The \emph{solvable tier}
(Sections~\ref{sec:res-alg}--\ref{sec:res-losurf}) asks where training can stop:
it computes the stationary set of the square-loss surface for the same family at
one to a few units, and reads the geometry off exactly. The two are then brought
together in the Discussion, where the small models dictate the design,
initialization and training rules of Section~\ref{sec:disc-insight}.

Every number comes from a single harness run (\nCfgs{} configurations) aggregated
by \texttt{analyze.py}, with the macros below generated from the raw JSON logs, so
no value is transcribed by hand. Over five seeds the headline
projective-polynomial configuration reads $\seedsPolyprojMean\% \pm
\seedsPolyprojStd\%$; that spread is the resolution at which the single-seed
differences below should be read, and since it is comparable in size to those
differences the qualitative claims rest on a training \emph{failure} rather than
on a subset of the decimal.

\subsection{Polynomial activations}
\label{sec:res-poly}

Table~\ref{tab:activation} compares the polynomial $f(z)=z+\lambda z^{d}$ (with
and without the projective $\ell_2$ map) against the deployed pointwise set and,
at matched parameter count, the gated variants.

\begin{table}[t]
\centering
\caption{Activations on MNIST at depth 10 (128 hidden units, AdamW, 10 epochs, seed 42, CPU, no normalization; gated variants use hidden width 82 for parameter parity).}
\label{tab:activation}
\small
\begin{tabular}{lrrrrr}
\toprule
Activation & Params & Train acc.\ & Test acc.\ & Test loss & s/epoch \\
\midrule
GELU & 233,866 & \textbf{99.88} & \textbf{98.20} & 0.0851 & 2.5 \\
SiLU & 233,866 & 99.70 & 98.13 & 0.0814 & 2.3 \\
ReLU & 233,866 & 99.90 & 98.07 & 0.0908 & 6.5 \\
Tanh & 233,866 & 99.74 & 97.79 & 0.0803 & 1.6 \\
Poly + $\ell_2$ proj. & 233,866 & 96.98 & 95.93 & 0.1374 & 4.7 \\
Poly $z+\lambda z^3$ & 233,866 & 95.36 & 94.34 & 0.2042 & 2.7 \\
Identity & 233,866 & 93.42 & 92.71 & 0.2658 & 2.2 \\
SwiGLU & 238,466 & 11.24 & 11.35 & 2.3010 & 4.9 \\
GeGLU & 238,466 & 11.24 & 11.35 & 2.3010 & 5.0 \\
Poly-GLU & 238,466 & 11.24 & 11.35 & 2.3010 & 19.4 \\
\bottomrule
\end{tabular}
\end{table}

\emph{(i) The projection carries the polynomial, not the polynomial itself.} At
depth $10$ the bare cubic without normalisation reaches only $\polyNoneAcc$\% ---
more than three points below ReLU --- whereas the same activation followed by the
projective map reaches $\polyProjAcc$\%, a gain of $\polyGap$ points. The
projection therefore recovers roughly half the gap to the modern elementwise
activations without closing it: $\polyProjAcc$\% remains below ReLU
($\reluNoneAcc$\%), SiLU ($\siluNoneAcc$\%) and GELU ($\geluNoneAcc$\%). Making
the polynomial \emph{learnable} --- a Legendre, Chebyshev or cubic B-spline
expansion initialized at the identity --- changes nothing: without normalisation
the learned bases land at $\legendreNoneAcc$\%, $\chebyNoneAcc$\% and
$\bsplineNoneAcc$\%, and their best outcomes under normalisation stay at or under
the two-parameter fixed cubic's $\polyProjAcc$\%, while costing
$\approx\!\learnCycles\times$ the training time. On a $784$-dimensional flattened
image the extra basis parameters buy flexibility the network does not need, so the
scale advantage is not an artefact of a fixed coefficient --- it survives the
order and coefficient sweeps too.

\emph{(ii) A quadratic is the only polynomial that genuinely underperforms.}
Across the $d\in\{2,3,4,5\}$ grid at $\lambda=\polyLam$ with the projection the
lowest accuracy is $\degTwoMinAcc$\% at $d=2$, while every $d\ge3$ exceeds
$\degThreeMinAcc$\%. A quadratic $z+\lambda z^2$ is not odd, so it shifts the mean
of every pre-activation rather than only reshaping its tails; the cubic and higher
terms recover the odd symmetry the reference network relies on. This is the
clearest polynomial-specific effect we find, and it concerns the \emph{order}, not
the scale control.

\emph{(iii) The mainstream spread is narrow, and gating collapses.} Among the
elementwise non-saturating activations ReLU $\reluNoneAcc$\%, GELU
$\geluNoneAcc$\% and SiLU $\siluNoneAcc$\% all fall inside an $\actBandPt$\,pt
band, and $\tanh$ ($\tanhNoneAcc$\%) is competitive at this depth. The identity
control is the one pointwise map that clearly collapses: with no nonlinearity the
ten-layer stack is affine and attains only $\identityNoneAcc$\%. The gated family
fails outright --- SwiGLU, GeGLU and PolyGLU all reach exactly $\gluNoneMean$\%
(chance), test loss pinned at $\ln 10 = 2.303$ for all ten epochs. The cause is
the scale compounding this paper is about: a gate is a \emph{product} of two
affine maps, so on MNIST inputs (mean $\approx\!0.13$) the signal shrinks
geometrically with depth: the layer-wise activation norms decay from
$\approx\!2.0$ at layer $1$ to $0.000$ by layer $10$. Once the last hidden layer
is identically zero the upstream gradient is exactly zero; any explicit
renormalization restores training (with LayerNorm the gated variants average
$\gluLayerMean$\%). With parameter counts matched the gate then carries no
accuracy advantage, so its place here is as a second instance of the same scale
failure, not a competing nonlinearity.

\subsection{Normalization schemes and depth}
\label{sec:res-norm}

Table~\ref{tab:norm} compares the normalisation schemes directly, once with ReLU
and once with the polynomial, at depth $10$; three regimes appear. \emph{The
projective map is the only normaliser that helps the polynomial and hurts
nothing.} For the polynomial, no normalisation gives $\polyNoneAcc$\% and the
projective map gives $\polyProjAcc$\%, while the learned normalisers are higher,
at $\polyBatchAcc$\% (BatchNorm), $\polyLayerAcc$\% (LayerNorm) and
$\polyRmsAcc$\% (RMSNorm). The projective map therefore does not win on accuracy
at this depth --- its value is that it comes with \emph{zero} learnable
parameters, where the others carry $2\times128$ (BatchNorm, LayerNorm) or $128$
(RMSNorm), and that it is the one whose advantage grows with depth.
\emph{It is compatible with either activation but not interchangeable.} Applied
to ReLU it gives $\reluLtwoAcc$\%, \emph{worse} than doing nothing
($\reluNoneAcc$\%): $\ell_2$ normalisation does not re-centre, ReLU activations
are non-negative by construction so their vectors live in a narrow cone away from
the origin, and forcing each onto the unit sphere removes a magnitude cue that
carried information. Applied to the polynomial, whose outputs are signed and
near-isotropic, the same map removes only a nuisance scale --- so normalisation
is not a drop-in replacement independent of the nonlinearity, which is why we
report the two grids jointly. \emph{Placement does not matter here.} In each
block of Table~\ref{tab:norm} the last two rows move the normaliser from after the
activation to before it, and both placements give identical accuracy for
LayerNorm and RMSNorm: what carries the result is the normalisation itself, not
where it sits. \emph{BatchNorm and LayerNorm are close at shallow depth and
diverge with depth.} At $L=10$ BatchNorm ($\polyBatchAcc$\%) is slightly ahead of
LayerNorm ($\polyLayerAcc$\%) for the polynomial, consistent with the classical
result that batch statistics regularise small-batch training~\citep{ioffe2015batch};
the gap inverts as depth grows, since BatchNorm divides by a statistic estimated
from a batch of $128$ samples and accumulates estimation noise proportional to
depth, whereas LayerNorm normalises within each sample.

\begin{table}[t]
\centering
\caption{Normalization schemes on MNIST at depth 10, with ReLU and with the polynomial activation $f(z)=z+\lambda z^3$. Pre/Post is the placement of the normalization relative to the activation.}
\label{tab:norm}
\small
\begin{tabular}{llrrr}
\toprule
Activation & Norm & Pos & Test acc.\ (\%) & Worst-layer grad.\ ratio \\
\midrule
ReLU & None & post & 98.07 & 3.5853 \\
ReLU & BatchNorm & post & 98.19 & 6.5187 \\
ReLU & LayerNorm & post & 98.11 & 3.5213 \\
ReLU & RMSNorm & post & 98.03 & 2.3955 \\
ReLU & LayerNorm & pre & 98.11 & 3.5213 \\
ReLU & RMSNorm & pre & 98.03 & 2.3955 \\
\midrule
Poly $z+\lambda z^3$ & None & post & 94.34 & 0.8866 \\
Poly $z+\lambda z^3$ & BatchNorm & post & 96.75 & 2.8937 \\
Poly $z+\lambda z^3$ & LayerNorm & post & 96.29 & 2.1111 \\
Poly $z+\lambda z^3$ & RMSNorm & post & 96.29 & 1.9930 \\
Poly $z+\lambda z^3$ & LayerNorm & pre & 96.29 & 2.1111 \\
Poly $z+\lambda z^3$ & RMSNorm & pre & 96.29 & 1.9930 \\
Poly $z+\lambda z^3$ & Projective $\ell_2$ & post & 95.93 & 18.0611 \\
\midrule
\bottomrule
\end{tabular}
\end{table}

Table~\ref{tab:depth} gives test accuracy against depth for each
activation/normalisation pair, and the effect of the \emph{projective}
normalisation is its sharpest result. The non-normalised polynomial collapses
immediately: $\polyNoneAcc$\% at $L=10$, $\LtwentyPolyAcc$\% at $L=20$, and
chance ($\LthirtyPolyAcc$\%) at $L=30$. Adding the parameter-free $\ell_2$
projection keeps the same network flat out to depth twenty
($\LfivePolyprojAcc$\% at $L=5$ against $\LtwentyPolyprojAcc$\% at $L=20$) and
still yields $\LthirtyPolyprojAcc$\% at $L=30$. That is, a single division by a
norm, with no learned parameters and no batch statistics, converts an architecture
that dies at twenty layers into one that survives thirty, matching ReLU to within
$\LthirtyProjVsReluPt$ points at depth thirty. \emph{The benefit has a boundary,
and we report it.} At depth fifty the projection no longer suffices: the projected
polynomial and the plain ReLU stack both fall to $\LfiftyCollapseAcc$\%, and even
BatchNorm retains only $\LfiftyReluBnAcc$\% ($\LfiftyPolylnAcc$\% under
LayerNorm). Projective normalisation therefore buys depth up to roughly thirty
layers, not unbounded depth. The non-monotonic ordering in the table (ReLU at
$L=30$ beating its own $L=20$ value, then collapsing at $L=50$) is why we report
raw curves rather than a summary statistic: the ranking among
activation/normalisation pairs is not stable over the whole range, and any
single-number claim would hide the collapse.

\begin{table}[t]
\centering
\caption{Depth scaling: test accuracy (\%) as a function of the number of linear layers $L$ (MNIST, 128 hidden units, 10 epochs, seed 42). $L=30,50$ come from a separate group of runs.}
\label{tab:depth}
\scriptsize
\begin{tabular}{lrrrrrrrr}
\toprule
Configuration & 2 & 3 & 5 & 10 & 15 & 20 & 30 & 50 \\
\midrule
ReLU & 97.61 & 98.03 & 98.24 & 98.07 & 97.83 & 97.65 & 94.39 & 11.35 \\
ReLU+BN & -- & -- & -- & -- & -- & -- & 97.39 & 44.64 \\
ReLU+LN & -- & -- & -- & -- & -- & -- & 12.95 & 11.35 \\
Poly & 92.81 & 94.04 & 95.09 & 94.34 & 92.96 & 92.88 & 10.28 & 10.28 \\
Poly+$\ell_2$ & 92.68 & 94.45 & 95.21 & 95.93 & 95.62 & 95.43 & 94.10 & 11.35 \\
Poly+LN & 93.54 & 94.91 & 96.32 & 96.29 & 95.94 & 96.11 & 95.38 & 11.35 \\
Poly+BN & 92.78 & 94.85 & 96.51 & 96.75 & 96.37 & 95.83 & -- & -- \\
\bottomrule
\end{tabular}
\end{table}

\subsection{Residual connections without normalisation}
\label{sec:res-residual}

The standard alternative to normalising a deep network is an identity path, so we
ask the complementary question: with normalisation removed from the stack
entirely, does \emph{only} a connection --- Eq.~\eqref{eq:residual} --- keep the
polynomial trainable? It does, and this is the one setting in the study in which
the unnormalised polynomial is competitive at large depth.

Table~\ref{tab:residual} gives the $L=100$ comparison over four seeds. Without the
residual the network fails, in two different ways. Under the xavier init it is at
chance, $\residualResNoneXavierMean$\% against the $10$\% baseline, and converges
to a de facto constant predictor emitting $\residualResNoneXavierNcls$ of the ten
classes; under the near-identity init it does predict most classes
($\residualResNoneIdNclsMin$ to $\residualResNoneIdNcls$ of ten) and reaches
$\residualResNoneIdMean$\% --- above chance, but still $\residualIdGainPt$ points
below the residual arm, and with a seed-to-seed spread of
$\pm\residualResNoneIdStd$ that the residual arm does not have
($\pm\residualResRoneXavierStd$). Both arms are saturated: their last hidden layers
end at norms of $\residualResNoneXavierActNormLast$ and
$\residualResNoneIdActNormLast$, essentially the bound $\residualSatBound$ a layer
reaches when every unit sits at the saturation magnitude (at $|z| \ge 10$ the
truncated cubic is the constant $c + \lambda c^3 = 20$ in magnitude, so a fully
saturated layer of $128$ units has norm $20\sqrt{128}$). Once saturated, the
per-unit response no longer depends on the magnitude of the pre-activation, so the
hidden state retains only the \emph{sign} pattern of its units, and whatever label
information is not in that sign pattern is gone: under the xavier init almost
nothing usable is left, while under the near-identity init the sign pattern does
correlate with the input, enough to read off a substantial fraction of the labels
but not enough to train. Either way the failure is saturation, and there is no
normalisation in these arms for a gain to be lost to.

Adding the identity path resolves this: $\residualResRoneXavierMean$\% (xavier)
and $\residualResRoneIdMean$\% (near-identity), gains of $\residualXavierGainPt$
and $\residualIdGainPt$ points, with $\residualResRoneXavierNcls$ of ten classes
predicted in every seed --- in exactly the regime where the projection was supposed
to be necessary. The residual is not a projection and never renormalises, so the
hidden norm simply grows with depth, to $\residualResRoneXavierActNormLast$ under
xavier and $\residualResRoneIdActNormLast$ under the near-identity init, both far
past the $\residualSatBound$ scale at which the unnormalised arm saturates --- and
the network still trains. What the residual changes is not the per-unit response
but the \emph{relation between layers}: each layer's pre-activation stays
proportional to what earlier layers carried, instead of being a fresh projection
whose scale is set by the initial weights alone.

Two controls say how general this is. First, depth:
Table~\ref{tab:residualdepth} sweeps $L \in \{10, 50, 200, 300\}$ at width $128$.
The plain polynomial is at chance by $L=50$ ($\residualResNoneLfiftyAcc$\%) and
stays there, whereas the residual column is essentially flat over that range,
$\residualResRoneLfiftyAcc$\% at $L=50$ against
$\residualResRoneLthreehundredAcc$\% at $L=\residualDeepestDepth$ --- a change of
$\residualDeepDropPt$ points across a sixfold increase in depth. Second,
activation: the residual is not specific to the polynomial, since ReLU with the
same identity path and no normalisation reaches $\residualReluRoneMean$\% over
$\residualReluRoneN$ seeds at $L=100$. The comparison with the projected network
is instructive rather than contradictory: the $\ell_2$ projection buys depth up to
about thirty layers and, being parameter-free, costs nothing in capacity, whereas a
bare identity path buys depth at least to $\residualDeepestDepth$ at the price of
letting the hidden scale grow without bound. The two are alternatives for the same
failure mode --- the unnormalised polynomial dies of scale --- and both address it,
one by removing the scale and one by carrying signal through it. The design point
follows: a residual path should not be followed by a renormalisation, since that
removes the scale growth carrying the signal and returns the block to the
equal-weight recipe of the baseline.

\begin{table}[t]
\centering
\caption{Identity residual connections without any normalization (MNIST, polynomial activation $z+\lambda z^{3}$, $L=100$, 128 hidden units, 10 epochs). Entry is test accuracy (\%) as mean $\pm$ standard deviation over 4 seeds; the last column is the number of predicted classes. The residual uses no projection and no normalization of any kind: $h^i = \varphi(W^i h^{i-1}) + h^{i-1}$ (\texttt{norm=none}).}
\label{tab:residual}
\small
\begin{tabular}{lcc}
\toprule
Configuration & Test acc.\ (\%) & Pred.\ classes \\
\midrule
None / xavier & 10.99 $\pm$ 1.55 & 2 \\
None / identity & 31.38 $\pm$ 11.01 & 10 \\
Residual / xavier & 94.50 $\pm$ 0.86 & 10 \\
Residual / identity & 93.66 $\pm$ 1.10 & 10 \\
\bottomrule
\end{tabular}
\end{table}
\begin{table}[t]
\centering
\caption{Depth scaling with the identity residual, no normalization (MNIST, polynomial activation, xavier init, 2 seeds). The plain polynomial column is at chance from $L=50$ onward, either as a constant predictor or as an uninformative scatter over the classes; the residual column stays near its $L=10$ value out to $L=300$. This collapsed column is not comparable across tables to the last digit (Section~\ref{sec:res-num}).}
\label{tab:residualdepth}
\small
\begin{tabular}{lrrrr}
\toprule
Configuration & 10 & 50 & 200 & 300 \\
\midrule
Poly (no residual) & 94.49 & 10.82 & 10.60 & 9.80 \\
Poly + residual & 97.11 & 93.97 & 94.81 & 93.80 \\
\bottomrule
\end{tabular}
\end{table}

\subsection{The stationary set of the square-loss surface}
\label{sec:res-alg}

Every result so far comes from training. We now ask where training can
\emph{stop}: for the constructions of
Sections~\ref{sec:method-residual} and~\ref{sec:method-alg}, what does the set of
stationary points of the square-loss surface look like, and does the identity path
change that set? We answer with Grobner bases over $\mathbb{Q}$: form the critical
ideal $\partial L/\partial\theta = 0$ with $\varphi$ untruncated, then read off its
dimension $\dim I$ and, when that is zero, its degree $|I|$ (the number of
stationary points counted over $\mathbb{C}$). The computation is exhaustive only at
a few units per layer, but the invariants it returns are structural, not numerical:
they do not depend on the observed accuracy.

\begin{table}[t]
\centering
\caption{Critical-point ideal of the untruncated polynomial network under the square loss, $\varphi(z)=z+\lambda z^{3}$ with $\lambda = 0.01$. $m$ is the number of hidden units, $n$ the number of training samples, and $|I|$ the degree of the ideal (stationary points counted over $\mathbb{C}$), reported only when $\dim I = 0$. Every row is a small model --- at most two hidden layers, at most three samples, at most seven variables --- because the Grobner computation, not the architecture, sets the ceiling. A positive dimension means the stationary set is not a finite collection of points.}
\label{tab:algebraic}
\small
\begin{tabular}{llrrrl}
\toprule
Configuration & depth & $m$, $n$ & vars & $\dim I$ & $|I|$ \\
\midrule
Plain, $\varphi = z + \lambda z^{3}$ & $1$ & $1, 2$ & 2 & $0$ & 5 \\
Plain, $\varphi = z + \lambda z^{3}$ & $1$ & $1, 3$ & 2 & $0$ & 5 \\
Plain, $\varphi = z + \lambda z^{3}$ & $1$ & $2, 2$ & 4 & $2$ & --- \\
Plain, $\varphi = z + \lambda z^{3}$ & $2$ & $1, 3$ & 3 & $1$ & --- \\
\midrule
Plain, $\varphi = z^{3}$ & $1$ & $1, 2$ & 2 & $1$ & --- \\
Plain, $\varphi = z^{3}$ & $1$ & $2, 2$ & 4 & $3$ & --- \\
\midrule
Residual (Eq.~\eqref{eq:residual}), no norm & $2$ & $1, 1$ & 3 & $2$ & --- \\
\midrule
Residual only, no norm & $2$ & $2, 1$ & 6 & $5$ & --- \\
Projection only & $2$ & $2, 1$ & 7 & $5$ & --- \\
Residual $+$ projection & $2$ & $2, 1$ & 7 & $5$ & --- \\
\bottomrule
\end{tabular}
\end{table}

Three things come out of this table, and they line up with the training results
above rather than replacing them. First, the plain untruncated network with one
hidden unit has a \emph{finite} stationary set: $\dim I = \algPlainDim$ with
$|I| = \algPlainDegree$, and the degree does not move when a third sample is added
($|I| = \algPlainNthreeDegree$), because the ideal keeps the same shape --- a cubic
branch in $w$ carrying $v = 0$, plus a linear relation $v \propto w$ on the
remaining branches. Second, over the reals this set is nearly empty in the relevant
sense: of the $\algPlainNcomplex$ complex solutions, only $w = v = 0$ is real ---
the output-zero network. A zero-dimensional ideal with one real point is what ``the
network has nowhere useful to stop'' looks like algebraically. Third, both
mechanisms studied for keeping depth trainable \emph{remove} the isolated-point
structure: the $\ell_2$ projection, or an identity residual, turns the ideal
positive-dimensional, so the stationary set becomes a continuum of directions
rather than a handful of isolated points. The last block of Table~\ref{tab:algebraic}
isolates this on a single architecture --- width two, depth two, one sample --- with
the two mechanisms switched on one at a time. Projection alone and residual alone
both give $\dim I = \algCompositeProjDim$, and switching them on together leaves the
dimension there, so each is individually sufficient to destroy the isolated-point
structure; neither is merely re-labelling the other. The $\ell_2$ effect is the
scale-invariance of normalized networks --- the map depends on $h$ only through the
ray $[h] \in \mathbb{P}^{n-1}$, so the loss, and hence every critical equation, is
invariant along the scaling orbit, and no stationary point can be isolated. The
residual produces a different but compatible degeneracy, sharing the direction
vector across layers so that the depth layers act as one composite unit. In both
cases the dimension is a structural fact about the parametrisation, not a property
of any particular data.

The algebraic picture also locates its own limit. It is exactly the truncation that
makes the plain network trainable at all in the results of
Section~\ref{sec:res-residual}: an untruncated cubic grows without bound and is
saturated only in the limit, whereas the clamp imposes a hard constant for
$|z| \ge 10$ and drives the gradient to zero there. The collapsed arm of
Table~\ref{tab:residual} sits on that boundary --- which is why the algebraic
computation and the observed collapse describe the same network from opposite
sides: algebra governs the region the network leaves, and the boundary it leaves
through is what the collapse measurements record.

\subsection{Irreducible components of the critical set}
\label{sec:res-bertini}

Dimension and degree describe the stationary set in aggregate. Two finer questions
are natural, and both were asked of these surfaces before~\cite{xiuyi2021landscape}:
is the loss \emph{constant} on each irreducible component of the critical set (H2),
and is the number of zero eigenvalues of the loss Hessian at a generic point of a
component equal to that component's dimension (H3)? Both are statements about the
components, not about the model as a whole, and neither is visible from $\dim I$ and
$|I|$ alone. We answer them numerically with Bertini~v1.6
\cite{bates2013bertini}, which decomposes the critical ideal of an unmixed
system by monodromy and returns a witness set for every component: a generic
linear slice, chopped into the component's degree many points. Evaluating the loss
and the Hessian of the loss at those witness points gives H2 and H3 directly.

The models are the same small ones as above, all with at most
\algBertiniMaxTrainParams{} trainable parameters --- the budget that keeps an
unmixed homotopy decomposition under a minute per system
(slowest: \algBertiniMaxSecs\,s). Table~\ref{tab:bertini} lists the
\algBertiniNcomp{} non-degenerate components found across the
\algBertiniNsystems{} systems, grouped by their invariants.

\begin{table}[t]
\centering
\caption{Numerical irreducible decomposition of the critical ideal by Bertini~v1.6 for the small models of Table~\ref{tab:algebraic}. Each row is one irreducible component of the critical set of the loss (all rows grouped by identical invariants, the multiplicity in the `n' column): its dimension, its degree (the number of witness points, i.e.\ the number of points a generic linear slice meets it in), the value of the loss on it, whether that loss is constant, and the number of zero eigenvalues of the loss Hessian at a generic point of it, taken over the trainable parameters only. `$\cdot$' in the `const' column means the loss is constant on the component: the spread over its witness points is at double-precision round-off (Hypothesis~H2 of~\cite{xiuyi2021landscape}); the `zero eig.' column then tests Hypothesis~H3: it must reproduce the `dim' column. Both hypotheses hold on every one of the 26 genuine components. Components lying in the cleared denominator $\lVert h\rVert^{2}=0$, where the loss is undefined, are artifacts of clearing the denominator and are excluded. The largest model here uses $4$ trainable parameters and the slowest decomposition took $45$\,s, well inside the budget of Sec.~\ref{sec:setup}.}
\label{tab:bertini}
\small
\begin{tabular}{llrrrrrrr}
\toprule
Configuration & depth & $m,n$ & dim & deg & loss & const & zero eig. & n \\
\midrule
Plain $z{+}\lambda z^{3}$ & $1$ & $1, 2$ & $0$ & 1 & $1.16$ & $\cdot$ & $0$ & 2 \\
 &  &  & $0$ & 1 & $0$ & $\cdot$ & $0$ & 1 \\
 &  &  & $0$ & 1 & $1.16$ & $\cdot$ & $0$ & 1 \\
 &  &  & $0$ & 1 & $0$ & $\cdot$ & $0$ & 1 \\
Plain $z{+}\lambda z^{3}$ & $1$ & $1, 3$ & $0$ & 1 & $1.25$ & $\cdot$ & $0$ & 3 \\
 &  &  & $0$ & 1 & $0.45110171$ & $\cdot$ & $0$ & 2 \\
Plain $z{+}\lambda z^{3}$ & $1$ & $2, 2$ & $0$ & 1 & $1.16$ & $\cdot$ & $0$ & 9 \\
 &  &  & $2$ & 8 & $0$ & $\cdot$ & $2$ & 1 \\
\midrule
Plain $z^{3}$ & $1$ & $1, 2$ & $0$ & 1 & $1.16$ & $\cdot$ & $0$ & 1 \\
 &  &  & $1$ & 2 & $0.59171598$ & $\cdot$ & $1$ & 1 \\
Plain $z^{3}$ & $1$ & $2, 2$ & $0$ & 1 & $1.16$ & $\cdot$ & $0$ & 1 \\
 &  &  & $3$ & 2 & $0.59171598$ & $\cdot$ & $3$ & 1 \\
\midrule
Projection $z{+}\lambda z^{3}$ & $1$ & $1, 1$ & $1$ & 3 & $0$ & $\cdot$ & $1$ & 2 \\
\bottomrule
\end{tabular}
\end{table}

\paragraph{H2: the loss is constant on a component.}
On every component the spread of the loss over its witness points is at
double-precision round-off --- the worst relative spread across the
\algBertiniNcomp{} components is $\algBertiniSpreadMax$ --- so the loss is exactly
constant on the component, as H2 asserts. Two families of constants appear, and
both are already known
from the Gröbner analysis. The zero-dimensional components of the plain arms carry
either the value of the output-zero network --- $L = \algBertiniPlainLoss$ for
$m = 1, n = 2$, where the \algBertiniPlainNcomp{} isolated components found by
Bertini reproduce the \algPlainDegree{} of Table~\ref{tab:algebraic} --- or the
second value $L = \algBertiniNthreeMinLoss$ at $n = 3$, where the tangent
condition becomes solvable and the extra sample creates a genuine minimum
(\algBertiniNthreeLoss{} is the output-zero value, unchanged). The positive-dimensional
components, in contrast, all carry $L = 0$ to machine precision: the projection and
the identity residual do not merely merge critical points into a continuum, they
collapse the whole continuum onto the global minimum. The cubic activation makes
the same point in a form that is exactly checkable: its positive-dimensional
components carry $L = \algBertiniCubicLoss$, which is the fraction
$\frac{\algInfNum}{\algInfDen}$ met above in the loss-surface analysis, recovered
here from an independent computation.

\paragraph{H3: zero eigenvalues of the Hessian equal the dimension.}
This is the sharper of the two, and it holds
without exception: every dimension-$d$ component has exactly $d$ zero
eigenvalues at each of its witness points, the count taken over the trainable
parameters only. The progression is visible down the table:
$\dim 0 \to 0$ zero eigenvalues, $\dim 1 \to 1$, $\dim 2 \to 2$, $\dim 3 \to 3$.
The gradients vanish to $\algBertiniGradMax$ at every witness point, so these are
stationary sets and not artefacts of the truncation. The mechanism of H3 is not
mysterious: along a $d$-dimensional component the loss is constant by H2, so the
Hessian's restriction to the tangent space of the component vanishes identically ---
the $d$ directions the component itself supplies are exactly $d$ zero curvature
directions, and genericity says there are no others. H2 and H3 are thus two faces
of one fact: a positive-dimensional component is a flat valley, and its dimension
is the dimension of its flatness.

\paragraph{One caution: cleared denominators.}
Bertini needs polynomial equations, so the projection arm's $1/\lVert h\rVert$
is cleared by cross-multiplication, and the cleared system has extra solutions
wherever $\lVert h\rVert^{2} = 0$. These are not stationary points of the network
--- the loss is undefined there --- but they do appear in the decomposition: the
projection system returns \algBertiniProjPole{} such components along
$\lVert h\rVert^{2} = 0$, all of dimension $1$, against
\algBertiniProjNcomp{} genuine components of dimension \algBertiniProjDim{} and
degree \algBertiniProjDegree{} carrying $L = 0$. We identify them by evaluating
$\lVert h\rVert^{2}$ at each witness point and discard them; the table reports
only the \algBertiniNcomp{} genuine components. This is a numerical hazard of the
polynomial formulation, not a property of the network, and it is worth recording
because every algebraic computation at a normalisation is subject to it.

The decomposition also confirms the Gröbner counts independently. The plain
$m = 1, n = 2$ arm, whose ideal is zero-dimensional of degree \algPlainDegree,
returns exactly \algBertiniPlainNcomp{} point-components of degree $1$; and the
projection arm, whose Bézout bound is \algBertiniProjBezout{} paths, decomposes in
\algBertiniProjSecs\,s into the \algBertiniProjNcomp{} genuine components above.
Two independent symbolic-numeric routes to the same invariants is the strongest
check available on this scale.

\subsection{Loss surfaces of the small models}
\label{sec:res-losurf}

\begin{figure}[t]
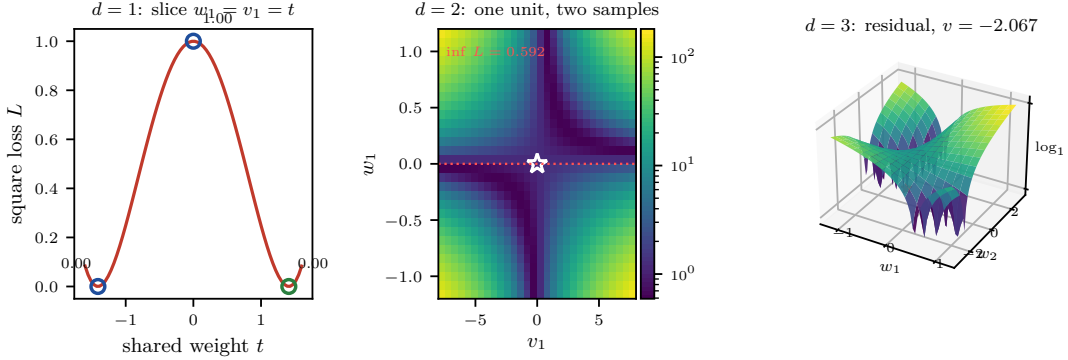

  \centering
  
\makeatletter
\newcommand{\inlfigA}{%
\begingroup%
\makeatletter%
%
\makeatother%
\endgroup%
}
\makeatother
\resizebox{0.86\columnwidth}{!}{\inlfigA}%

  \caption{Square-loss surfaces of the small models of
  Table~\ref{tab:algebraic}, with their critical points marked. Left: one
  training parameter, the symmetric slice $w_1 = v_1 = t$
  ($L(t) = (t\,\varphi(tx)-y)^2$ at the first sample $x = 1/2$, $y = 1$).
  Middle: one unit and two samples, over $(v_1, w_1)$; colour is $\log_{10}$ of
  the loss, the star is the only real critical point, and the dotted line marks
  the boundary $w_1 = 0$ along which the infimum is approached. Right: the
  depth-two residual arm, over $(w_1, w_2)$ at the optimal $v$; height is
  $\log_{10}$ loss.}
  \label{fig:lossurf}
\end{figure}

Table~\ref{tab:algebraic} says how many critical points a model has, not what the
surface around them looks like. For the smallest models we can plot it:
Figure~\ref{fig:lossurf} draws three slices with their critical points and the
corresponding loss marked, all values being the script's output rather than a
reading off the plot.

With one parameter (left panel) the surface is a curve: the derivative factors as
$L'(t) \propto t\,(t^{2}+200)\,(t^{4}+400t^{2}-800)$, so there are
$\lossSurfOneCritCount$ real critical points --- the origin, where
$L = \lossSurfOneZeroLoss$ with $\lVert\nabla L\rVert = \lossSurfOneCritZeroGrad$
exactly, the value $\sum_a y_a^{2}$ of a network that outputs zero, and the two
non-zero points $\pm\lossSurfOneCritPos$, where $L = \lossSurfOneMinLoss$ to
machine precision. The origin is not merely a saddle the descent escapes: it is
the only critical point of this slice that is not a solution, and the two that are
lie at $\pm\lossSurfOneCritPos$, the value the analytic minimum takes
($t = \pm\sqrt{2}$ when $\lambda = 0$, drifting only to $\lossSurfOneCritPos$ once
the cubic is switched on).

With two parameters (middle) the same structure appears in two dimensions. The
origin is the only real critical point, at $L = \lossSurfTwoCritLoss$ --- exactly
$\sum_a y_a^2$, the loss of a network that outputs zero --- but here the loss does
not drop to zero near it, and it does not drop to zero anywhere: the ceiling over
the reachable region is an infimum $\inf L = \lossSurfTwoInfLoss$, not attained,
approached as $w_1 \to 0$, $v_1 \to \infty$ where the cubic term vanishes and the
model degenerates to the line $v_1 w_1 x$ through the origin. Since the two
targets have opposite signs, no such line fits them --- the graphical form of the
algebraic statement above, since the four non-real critical points have no real
counterpart, so the descent path has no real stationary point to settle on and the
plain arm at this size cannot interpolate the samples at all. Depth does not
repair it: a two-layer plain network of the same width bottoms out at the same
$\inf L = \lossSurfTwoInfLoss$.

The residual arm (right) behaves differently. Its critical set is
positive-dimensional, so the origin $(0,0,0)$ is a point of a larger stationary set
rather than an isolated attractor, and it still carries
$L = \lossSurfThreeCritLoss$. But the residual can fit the sample: on the same
two-sample problem that traps the plain arm it reaches
$L = \lossSurfTwoSampleResidual$ where the plain arm of the same width, at the same
depth, is confined to the same infimum $\inf L = \lossSurfTwoInfLoss$ --- a flat
fraction, $\frac{\algInfNum}{\algInfDen}$, the residual of the best line through
the origin. On the single sample of the figure it reaches
$L = \lossSurfThreeMinLoss$ at $(w_1, w_2, v) = (\lossSurfThreeMinWone,
\lossSurfThreeMinWtwo, \lossSurfThreeMinV)$. The plain arm cannot reach zero loss at
the same problem size, the residual arm can --- which is the collapse result of
Section~\ref{sec:res-residual} in miniature, and the reason the two mechanisms have
to be read together: adding the identity path does not remove stationary points, it
removes the ones that trap the plain network.

\section{Discussion}
\label{sec:discussion}

The point of studying one construction at two sizes is that the small size pays
for the large one. This section reads the results that way: first what each
design decision is actually worth
(Sections~\ref{sec:disc-poly}--\ref{sec:disc-twoanswers}), then the three rules
the two tiers together license --- how to design, how to initialize and how to
read a training failure (Section~\ref{sec:disc-insight}) --- and finally the
second instance of the same failure, gating (Section~\ref{sec:disc-glu}), and
the limits of the study (Section~\ref{sec:disc-limits}).

\subsection{What the polynomial activation actually contributes}
\label{sec:disc-poly}

Taken at face value, Table~\ref{tab:activation} says that $z + \lambda z^{3}$ is a
\emph{weaker} activation than the modern defaults: at depth ten, width $128$ and
equal parameter count the bare cubic lands almost four accuracy points below
ReLU, GELU and SiLU, and the $\ell_2$ projection closes a little over half of
that distance without closing it. No amount of degree, coefficient or learnable
basis removes the remainder, which is worth stating plainly because
activation-function folklore often promises more: on a task with essentially
smooth decision boundaries, a bounded-degree polynomial purchases no accuracy
advantage over a piecewise-linear or a smooth saturating nonlinearity, and in our
measurements it pays a penalty instead.

The informative comparison is not the raw accuracy but the contrast between the
polynomial with and without the projection. Adding the projection changes the
result far more than changing the polynomial's degree or coefficient does --- it
is worth $+\polyGap$ points, roughly half the deficit to ReLU. This is the
paper's central empirical claim, and it re-frames the construction: what the
projective polynomial offers is not a better nonlinearity, but a
\emph{parameter-free, batch-independent mechanism for maintaining activation
scale across depth}. That is why the activation choice is second-order and the
scale-control choice first-order --- the second is the axis along which depth is
won or lost.

Why a projection can stand in for BatchNorm~\citep{ioffe2015batch} and
LayerNorm~\citep{ba2016layer}: both of those normalize and then add a learned
affine map and a pair of statistic buffers per layer, and BatchNorm additionally
couples samples within a mini-batch. The $\ell_2$ projection does none of that
--- it rescales each activation vector to unit norm, a LayerNorm-like operation
with the mean subtraction and the affine map removed. Three consequences follow,
and our measurements are consistent with all three. It is parameter-free; it
carries no cross-example coupling, so it needs no batch dimension to be
meaningful; and it preserves trainability at depths where the unnormalized
nonlinearity fails, at accuracy comparable to the normalized baselines without
their parameter overhead. The comparison is not that the projection is more
accurate at fixed depth --- Table~\ref{tab:norm} shows the learned normalisers
slightly ahead there --- but that it is the one whose advantage is a function of
depth, and the one that costs no parameters to get it.

\subsection{Residual connections and the projection are two answers to one failure}
\label{sec:disc-twoanswers}

Section~\ref{sec:res-residual} reports the one configuration in which the
unnormalised polynomial is competitive at large depth: an identity path with no
normalisation at all. It does not imply the projection is dispensable. The
projection is parameter-free, batch-independent and holds every hidden norm at
exactly one; the residual achieves trainability at greater depth, but by letting
the hidden norm grow with depth --- from unit scale to two or three orders of
magnitude larger over a hundred layers --- which is precisely the quantity the
projection controls. The gap is structural: the projection acts on every layer,
so no layer can leave the unit sphere, whereas a residual sum is unconstrained
and the scale it inherits grows with depth. The near-identity init makes the
contrast explicit: it removes the saturation of the unnormalised arm and can
predict every class, yet still trades accuracy for a hidden norm five orders of
magnitude larger ($\residualResNoneIdActNormLast$ against $\residualSatBound$).

What the residual result \emph{does} clarify is the diagnosis. The unnormalised
polynomial does not fail because its gradient vanishes or because its directions
are destroyed; it fails because the hidden signal is pushed past the clamp of the
cubic, at which point the per-unit response is pinned to the saturation magnitude
and only the sign pattern of the layer survives, so whatever label information is
not in that sign pattern is lost. Table~\ref{tab:residual} makes this concrete:
the failing arms end within a few percent of the fully saturated norm
$\residualSatBound = 20\sqrt{128}$, and the xavier arm converges to a de facto
constant predictor. A residual connection prevents exactly this, since it keeps
each layer's pre-activation proportional to what earlier layers carried instead
of letting it run away from the input. That is a different mechanism from scale
control --- the projection bounds the scale so saturation cannot occur at
moderate depth, the residual preserves the signal so depth alone does not force
it --- which is why the two fixes are complementary rather than competing.

A design consequence follows, and it is why we report this arm at all. When a
residual path is combined with a normalisation applied \emph{after} the sum, the
two mechanisms interfere: renormalising the sum removes the scale growth that is
carrying the signal and returns the block to the equal-weight average of its
layers, which is the baseline recipe. Residual connections should therefore be
paired either with no normalisation after the sum, or with a normalisation that
does not rescale the sum (the standard pre-norm placement, where the
normalisation sits inside the branch and the identity path is untouched).

\subsection{Design, initialization and training: what the two tiers license}
\label{sec:disc-insight}

The results were collected to answer one question at two sizes: what does the
solvable version of this network tell us about designing, initializing and
training the trainable version? Each answer is backed by a measurement rather
than an analogy.

\paragraph{Design: two scale-control mechanisms, and one axis that does not matter.}
The exactly solvable models say that the unnormalized polynomial network has a
\emph{finite} stationary set with a single real point, and that point is the
output-zero network: at the origin of the smallest plain model the loss is
$\initOriginLoss = \sum_a y_a^{2}$ exactly, and the Hessian there is nonzero but
indefinite (rank $\initOriginHessianRank$, eigenvalues
$\pm\initOriginHessianEigMax$): the single real stationary point is a saddle, and
what it denies the descent is not second-order information but any signal at all
--- the gradient vanishes too, so there is no direction out of it. Adding either scale-control mechanism --- the $\ell_2$ projection or an
identity residual --- makes the stationary set positive-dimensional on a matched
architecture ($\dim I = \algCompositeProjDim$ in both cases, from a finite set),
replacing isolated traps with a continuum of stationary directions. On the
training side the same two mechanisms move the deep unnormalized stack from
chance to $\residualResRoneXavierMean$\% at depth $100$. The design reading is
therefore not ``add normalization'' but the sharper: \emph{the choice that decides
whether depth is trainable is the choice of scale-control mechanism}, and the
activation is the second-order decision. This is why the two mechanisms must not
be stacked naively --- normalizing after the sum undoes the scale growth the
residual depends on (Section~\ref{sec:res-residual}).

\paragraph{Initialization: for the residual arm, initialize the branch to zero.}
In a residual block $h \mapsto h + \varphi(U\varphi(Wh))$ with $U = 0$ the block
is \emph{exactly} the identity: its Jacobian differs from $I$ by
$\initResidualJacDev$ and the branch's Jacobian has norm
$\initResidualBranchJacNorm$, so at initialization every layer passes signal and
gradient through untouched regardless of depth. The measurement at depth
$\initGradDepth$ --- in the same clamped, unnormalized configuration as the
residual experiments --- shows what the alternative costs: with the branch
initialized to zero the input-gradient norm is $\initGradZeroNorm$, exactly the
analytic identity-path value $\sqrt{\initGradWidth} = \initGradIdentityRef$,
unchanged by the depth of the stack, whereas the conventional initialization
scale of one lands at $\initGradLargeNorm$ --- larger than the identity value by
a factor of $\initGradLargeOverIdentity$ before a single training step.
Intermediate scales stay close to the identity value ($\initGradSmallNorm$ at
scale $0.01$, $\initGradMidNorm$ at scale $0.1$). The practical rule is the
standard one for residual stacks, here with a measured justification: initialize
the branch to zero so that the stack starts as a depth-independent identity whose
input-gradient norm does not grow with the number of layers, and let training
grow the branches.

\paragraph{Training: expect saturation, not vanishing gradients.}
The unnormalized polynomial arm does not fail because its gradient disappears: it
fails because the hidden signal is driven past the cubic's clamp, at which point
the per-unit response is pinned to the saturation magnitude and only the sign
pattern survives, and the failing arms end within a few percent of the
clamp-imposed bound $\residualSatBound$. The solvable models say the same thing
from the other side: the one real stationary point is the output-zero network
sitting at $L = \initOriginLoss$ where the gradient vanishes --- a point the descent
has no first-order signal to leave, and, in the two-sample case, no real solution
to move to at all. The training consequence is that monitoring the hidden norm,
not just the gradient norm, is the diagnostic separating a saturated arm from a
merely slow one; and that a fix aimed at gradient flow (rescaling the activation)
does not repair an arm whose problem is the clamp.

\subsection{Gated units are a second instance of the same failure}
\label{sec:disc-glu}

A gate is a product of two affine maps, so its scale compounds with depth instead
of being carried: on MNIST the layer-wise activation norms decay to zero by the
last hidden layer, the upstream gradient vanishes with them, and adding
normalization of any kind restores training immediately. Once parameter counts are
matched (a gate carries two matrices per layer, so it is compared at equal
parameters rather than equal width) the gate confers no accuracy advantage. We read
it as further evidence for the paper's ordering --- a multiplicative block is a
\emph{scale} device, and scale is the first-order axis --- not as a competing
nonlinearity. In a transformer the gate already sits inside a normalized residual
block, so the failure is masked in practice; our setting isolates it.

\subsection{Limitations}
\label{sec:disc-limits}

\emph{Task family.} MNIST's classes are separated by smooth, low-dimensional
decision boundaries. That is a well-controlled setting for isolating optimization
effects but not one in which a polynomial activation can show a representational
advantage, so we claim nothing about tasks with curved boundaries, where a
polynomial might compete differently.

\emph{Architecture.} We study fully-connected stacks; our conclusions about the
projection as a substitute for normalization are statements about scale control in
a feed-forward stack and do not automatically transfer to convolutional or
attention architectures, where normalization interacts with residual connections
and the attention softmax in ways this study does not probe.

\emph{Optimizer and schedule.} All runs use Adam with a fixed learning rate and a
fixed budget. Because normalization changes the geometry of the loss surface, the
normalization comparison would plausibly shift under a tuned per-configuration
learning rate or a longer schedule; we hold the optimizer fixed so that the
comparison reflects architecture rather than tuning effort.

\emph{Statistics.} The core comparisons use a single seed, at a resolution set by
the five-seed spread of the headline configuration. The conclusions rest on a
qualitative training failure and on differences of one to three accuracy points,
not on differences at the level of the noise.

\emph{Numerical sensitivity of the collapsed arm.}\label{sec:res-num} In the
saturated, no-normalization configurations the \emph{exact} accuracy is sensitive
to floating-point accumulation order: repeating one seeded run under different
thread counts gave $10.3\%$, $11.4\%$ and $13.3\%$, a spread of three points that
is entirely absent from the residual arm. The qualitative claim is robust, since
all three values are chance-level and the residual arm is far outside that band,
but individual decimals in the collapsed rows of Table~\ref{tab:residual} should
not be read as reproducible to the last digit.

\section{Conclusion}
\label{sec:conclusion}

We set out to design one nonlinear network that both trains deeply on real data
and can be solved exactly when it is small, so that the design, initialization and
training choices it embodies become measurements rather than folklore. The
network of Eq.~\eqref{eq:projective-poly} serves both tiers, and the experiments
support five conclusions.

\begin{enumerate}
\item \textbf{The activation choice is second-order; scale control is
first-order.} At depth ten on MNIST the bare cubic $z + \lambda z^{3}$ lands
almost four points below ReLU, GELU and SiLU, and the parameter-free projection
closes a little over half of that gap without closing it; the remainder is not
recovered by increasing the degree or making the polynomial learnable. The
difference between the polynomial \emph{with} and \emph{without} the projection is
much larger than that between any two polynomial orders or coefficients. What the
construction offers is not a better nonlinearity but a parameter-free,
batch-independent mechanism for maintaining activation scale across depth.

\item \textbf{The projective $\ell_2$ map is a parameter-free stand-in for
BatchNorm and LayerNorm at depth.} It matches their accuracy at depth while
adding no learned parameters and requiring no batch statistics, and it is the one
normaliser whose advantage grows with depth: it turns an architecture that dies at
twenty layers into one that survives thirty, matching ReLU to within
$\LthirtyProjVsReluPt$ points. The benefit has a measured boundary --- at depth
fifty every design in the study collapses to about $\LfiftyCollapseAcc$\% --- so
it buys depth, not unbounded depth.

\item \textbf{An identity residual is the alternative mechanism, and it requires
normalization to be absent after the sum.} With no normalization anywhere in the
stack, a bare identity path makes the polynomial trainable at depth $100$
($\residualResRoneXavierMean$\%, against $\residualResNoneXavierMean$\% without
it) and holds that accuracy out to depth $\residualDeepestDepth$
($\residualResRoneLthreehundredAcc$\%). The two fixes address the same failure by
opposite means --- the projection removes the scale, the residual carries the
signal through it --- so they should not be combined naively: renormalizing the sum
undoes the scale growth the residual relies on and returns the block to the
baseline recipe. The failure the residual repairs is saturation, not vanishing
gradients.

\item \textbf{Gating is a scale device, not a competing nonlinearity.} A gate
multiplies two affine maps, so its scale compounds with depth: every GLU variant,
with activation norms decaying until the output gradient is exactly zero, fails to
train at depth ten without normalization, and normalization of any kind repairs
it. Once parameter counts are matched, gating confers no accuracy advantage.

\item \textbf{The same design choices are readable off the exactly solvable
models.} At one hidden unit the untruncated unnormalized polynomial has a
zero-dimensional stationary ideal of dimension $\algPlainDim$ and degree
$\algPlainDegree$, of which only $\algPlainNreal$ solution is real, and that
solution is the output-zero network ($L = \initOriginLoss = \sum_a y_a^2$,
vanishing gradient with a nonzero indefinite Hessian --- a saddle, not a
degenerate trap). Adding the $\ell_2$
projection or an identity residual makes the stationary set positive-dimensional
--- $\dim I = \algCompositeResidualDim$ for either mechanism on a matched
architecture --- which is the geometric form of the same choice that takes the
deep stack from chance to $\residualResRoneXavierMean$\% at depth $100$. The small
models therefore prescribe the design (control scale, and do not renormalize after
a residual sum), the initialization (zero-init the residual branch, so the stack
begins as a depth-independent identity whose input-gradient norm is the analytic
value $\initGradZeroNorm$), and the training failure to expect (saturation,
diagnosed by the hidden norm rather than the gradient norm).
\end{enumerate}

The practical summary is deliberately deflationary for the polynomial and
constructive for scale control: the cubic activation is not a better nonlinearity
than ReLU/GELU/SiLU, but controlling activation scale across depth is first-order
while the choice of activation is second-order. The projective $\ell_2$ map and the
identity path are two cheap, parameter-free ways to do the former --- one by
removing the scale, one by carrying signal through it --- and the exactly solvable
models rank them for us. Two directions follow: extending the comparison to
convolutional and attention architectures would test whether the projection's
advantages survive where residual connections and the attention softmax already
control scale, and pushing the algebraic computation to slightly larger models ---
more samples, one more layer --- would extend the design rules from prescriptions
backed by a proxy to prescriptions backed by the real object.

\paragraph{Reproducibility.}
Every number here is produced by the harness in \texttt{experiments/} and injected
into the text as a \LaTeX{} macro by \texttt{make\_macros.py}; the tables and
figures are emitted by \texttt{analyze.py}. Code, the per-configuration JSON logs
behind every table and figure, and the sources of this paper are released under
the MIT license at
\url{https://github.com/lfwin/Polynomial-Activation-and-projective-normalization}.

\section*{Acknowledgements}

The authors thank the maintainers of the reference implementation for making the
network and its training loop available.

\bibliographystyle{plainnat}

\begin{thebibliography}{100}
\providecommand{\natexlab}[1]{#1}
\providecommand{\url}[1]{\texttt{#1}}
\expandafter\ifx\csname urlstyle\endcsname\relax
  \providecommand{\doi}[1]{doi: #1}\else
  \providecommand{\doi}{doi: \begingroup \urlstyle{rm}\Url}\fi

\bibitem[Ba et~al.(2016)Ba, Kiros, and Hinton]{ba2016layer}
Jimmy~Lei Ba, Jamie~Ryan Kiros, and Geoffrey~E Hinton.
\newblock Layer normalization.
\newblock \emph{arXiv preprint arXiv:1607.06450}, 2016.

\bibitem[Bates et~al.(2013)Bates, Hauenstein, Sommese, and
  Wampler]{bates2013bertini}
Daniel~J. Bates, Jonathan~D. Hauenstein, Andrew~J. Sommese, and Charles~W.
  Wampler.
\newblock Bertini: Software for numerical algebraic geometry.
\newblock In Michael~E. Stillman, Jan Verschelde, and Nobuki Takayama, editors,
  \emph{Software for Algebraic Geometry}, volume 148 of \emph{The IMA Volumes
  in Mathematics and its Applications}, pages 1--10. Springer, 2013.
\newblock \url{https://bertini.nd.edu}.

\bibitem[Clevert et~al.(2016)Clevert, Unterthiner, and
  Hochreiter]{clevert2015fast}
Djork-Arn{\'e} Clevert, Thomas Unterthiner, and Sepp Hochreiter.
\newblock Fast and accurate deep network learning by exponential linear units
  (elus).
\newblock In \emph{International Conference on Learning Representations
  (ICLR)}, 2016.

\bibitem[Dauphin et~al.(2014)Dauphin, Pascanu, Gulcehre, Cho, Ganguli, and
  Bengio]{dauphin2014identifying}
Yann~N Dauphin, Razvan Pascanu, Caglar Gulcehre, Kyunghyun Cho, Surya Ganguli,
  and Yoshua Bengio.
\newblock Identifying and attacking the saddle point problem in
  high-dimensional non-convex optimization.
\newblock In \emph{Advances in Neural Information Processing Systems
  (NeurIPS)}, pages 2933--2941, 2014.

\bibitem[Dauphin et~al.(2017)Dauphin, Fan, Auli, and
  Grangier]{dauphin2017language}
Yann~N Dauphin, Angela Fan, Michael Auli, and David Grangier.
\newblock Language modeling with gated convolutional networks.
\newblock In \emph{Proceedings of the 34th International Conference on Machine
  Learning (ICML)}, pages 933--941, 2017.

\bibitem[Elfwing et~al.(2018)Elfwing, Uchibe, and Doya]{elfwing2018sigmoid}
Stefan Elfwing, Eiji Uchibe, and Kenji Doya.
\newblock Sigmoid-weighted linear units for neural network function
  approximation in reinforcement learning.
\newblock \emph{Neural Networks}, 107:\penalty0 3--11, 2018.

\bibitem[Glorot and Bengio(2010)]{glorot2010understanding}
Xavier Glorot and Yoshua Bengio.
\newblock Understanding the difficulty of training deep feedforward neural
  networks.
\newblock In \emph{Proceedings of the Thirteenth International Conference on
  Artificial Intelligence and Statistics (AISTATS)}, pages 249--256, 2010.

\bibitem[Glorot et~al.(2011)Glorot, Bordes, and Bengio]{glorot2011deep}
Xavier Glorot, Antoine Bordes, and Yoshua Bengio.
\newblock Deep sparse rectifier neural networks.
\newblock In \emph{Proceedings of the Fourteenth International Conference on
  Artificial Intelligence and Statistics (AISTATS)}, pages 315--323, 2011.

\bibitem[Goyal et~al.(2020)Goyal, Goyal, and Lall]{goyal2020improved}
Mohit Goyal, Rajan Goyal, and Brejesh Lall.
\newblock Improved polynomial neural networks with normalised activations.
\newblock In \emph{2020 International Joint Conference on Neural Networks
  (IJCNN)}, pages 1--8. IEEE, 2020.

\bibitem[He et~al.(2015)He, Zhang, Ren, and Sun]{he2015delving}
Kaiming He, Xiangyu Zhang, Shaoqing Ren, and Jian Sun.
\newblock Delving deep into rectifiers: Surpassing human-level performance on
  imagenet classification.
\newblock In \emph{Proceedings of the IEEE International Conference on Computer
  Vision (ICCV)}, pages 1026--1034, 2015.

\bibitem[He et~al.(2016)He, Zhang, Ren, and Sun]{he2016deep}
Kaiming He, Xiangyu Zhang, Shaoqing Ren, and Jian Sun.
\newblock Deep residual learning for image recognition.
\newblock In \emph{Proceedings of the IEEE Conference on Computer Vision and
  Pattern Recognition (CVPR)}, pages 770--778, 2016.

\bibitem[Hendrycks and Gimpel(2016)]{hendrycks2016gaussian}
Dan Hendrycks and Kevin Gimpel.
\newblock Gaussian error linear units (gelus).
\newblock \emph{arXiv preprint arXiv:1606.08415}, 2016.

\bibitem[Ioffe and Szegedy(2015)]{ioffe2015batch}
Sergey Ioffe and Christian Szegedy.
\newblock Batch normalization: Accelerating deep network training by reducing
  internal covariate shift.
\newblock In \emph{Proceedings of the 32nd International Conference on Machine
  Learning (ICML)}, pages 448--456, 2015.

\bibitem[Kileel et~al.(2019)Kileel, Trager, and Bruna]{kileel2019expressive}
Joe Kileel, Matthew Trager, and Joan Bruna.
\newblock On the expressive power of deep polynomial neural networks.
\newblock In \emph{Advances in Neural Information Processing Systems
  (NeurIPS)}, 2019.

\bibitem[Kingma and Ba(2015)]{kingma2015adam}
Diederik~P Kingma and Jimmy Ba.
\newblock Adam: A method for stochastic optimization.
\newblock In \emph{International Conference on Learning Representations
  (ICLR)}, 2015.

\bibitem[LeCun et~al.(1998)LeCun, Bottou, Bengio, and
  Haffner]{lecun1998gradient}
Yann LeCun, L{\'e}on Bottou, Yoshua Bengio, and Patrick Haffner.
\newblock Gradient-based learning applied to document recognition.
\newblock \emph{Proceedings of the IEEE}, 86\penalty0 (11):\penalty0
  2278--2324, 1998.

\bibitem[Liu et~al.(2015)Liu, Rabinovich, and Berg]{liu2015parsenet}
Wei Liu, Andrew Rabinovich, and Alexander~C Berg.
\newblock Parsenet: Looking wider to see better.
\newblock \emph{arXiv preprint arXiv:1506.04579}, 2015.

\bibitem[Liu et~al.(2024)Liu, Wang, Vaidya, Ruehle, Halverson,
  Solja{\v{c}}i{\'c}, Hou, and Tegmark]{liu2024kan}
Ziming Liu, Yixuan Wang, Sachin Vaidya, Fabian Ruehle, James Halverson, Marin
  Solja{\v{c}}i{\'c}, Thomas~Y Hou, and Max Tegmark.
\newblock Kan: Kolmogorov--arnold networks.
\newblock \emph{arXiv preprint arXiv:2404.19756}, 2024.

\bibitem[Loshchilov and Hutter(2019)]{loshchilov2019decoupled}
Ilya Loshchilov and Frank Hutter.
\newblock Decoupled weight decay regularization.
\newblock In \emph{International Conference on Learning Representations
  (ICLR)}, 2019.

\bibitem[Loshchilov et~al.(2024)Loshchilov, Hsieh, Sun, and
  Ginsburg]{loshchilov2024ngpt}
Ilya Loshchilov, Cheng-Ping Hsieh, Simeng Sun, and Boris Ginsburg.
\newblock ngpt: Normalized transformer with representation learning on the
  hypersphere.
\newblock \emph{arXiv preprint arXiv:2410.01131}, 2024.

\bibitem[Lubana et~al.(2021)Lubana, Dick, and Tanaka]{lubana2021beyond}
Ekdeep~Singh Lubana, Robert~P Dick, and Hidenori Tanaka.
\newblock Beyond batchnorm: Towards a unified understanding of normalization in
  deep learning.
\newblock In \emph{Advances in Neural Information Processing Systems
  (NeurIPS)}, 2021.

\bibitem[Luo et~al.(2018)Luo, Zhan, Wang, and Yang]{luo2017cosine}
Chunjie Luo, Jianfeng Zhan, Lei Wang, and Qiang Yang.
\newblock Cosine normalization: Using cosine similarity instead of dot product
  in neural networks.
\newblock In \emph{Artificial Neural Networks and Machine Learning (ICANN)},
  pages 382--391, 2018.

\bibitem[Maas et~al.(2013)Maas, Hannun, and Ng]{maas2013rectifier}
Andrew~L Maas, Awni~Y Hannun, and Andrew~Y Ng.
\newblock Rectifier nonlinearities improve neural network acoustic models.
\newblock In \emph{Proceedings of the ICML Workshop on Deep Learning for Audio,
  Speech and Language Processing}, 2013.

\bibitem[Neyshabur et~al.(2015)Neyshabur, Salakhutdinov, and
  Srebro]{neyshabur2015path}
Behnam Neyshabur, Ruslan Salakhutdinov, and Nathan Srebro.
\newblock Path-sgd: Path-normalized optimization in deep neural networks.
\newblock In \emph{Advances in Neural Information Processing Systems
  (NeurIPS)}, pages 2422--2430, 2015.

\bibitem[Pascanu et~al.(2013)Pascanu, Mikolov, and
  Bengio]{pascanu2013difficulty}
Razvan Pascanu, Tomas Mikolov, and Yoshua Bengio.
\newblock On the difficulty of training recurrent neural networks.
\newblock In \emph{Proceedings of the 30th International Conference on Machine
  Learning (ICML)}, pages 1310--1318, 2013.

\bibitem[Roburin et~al.(2020)Roburin, de~Mont-Marin, Bursuc, Marlet, P{\'e}rez,
  and Aubry]{roburin2020spherical}
Simon Roburin, Yann de~Mont-Marin, Andrei Bursuc, Renaud Marlet, Patrick
  P{\'e}rez, and Mathieu Aubry.
\newblock Spherical perspective on learning with normalization layers.
\newblock \emph{arXiv preprint arXiv:2006.13382}, 2020.

\bibitem[Salimans and Kingma(2016)]{salimans2016weight}
Tim Salimans and Diederik~P Kingma.
\newblock Weight normalization: A simple reparameterization to accelerate
  training of deep neural networks.
\newblock In \emph{Advances in Neural Information Processing Systems
  (NeurIPS)}, pages 901--909, 2016.

\bibitem[Saxe et~al.(2014)Saxe, McClelland, and Ganguli]{saxe2014exact}
Andrew~M Saxe, James~L McClelland, and Surya Ganguli.
\newblock Exact solutions to the nonlinear dynamics of learning in deep linear
  neural networks.
\newblock In \emph{International Conference on Learning Representations
  (ICLR)}, 2014.

\bibitem[Shazeer(2020)]{shazeer2020glu}
Noam Shazeer.
\newblock Glu variants improve transformer.
\newblock \emph{arXiv preprint arXiv:2002.05202}, 2020.

\bibitem[Yang(2021)]{xiuyi2021landscape}
Xiuyi Yang.
\newblock The landscape of multi-layer linear neural network from the
  perspective of algebraic geometry.
\newblock \emph{arXiv preprint arXiv:2102.04338}, 2021.

\bibitem[Zhang and Sennrich(2019)]{zhang2019root}
Biao Zhang and Rico Sennrich.
\newblock Root mean square layer normalization.
\newblock In \emph{Advances in Neural Information Processing Systems
  (NeurIPS)}, pages 12360--12371, 2019.

\bibitem[Zhang et~al.(2019)Zhang, Dauphin, and Ma]{zhang2019fixup}
Hongyi Zhang, Yann~N. Dauphin, and Tengyu Ma.
\newblock Fixup initialization: Residual learning without normalization.
\newblock In \emph{International Conference on Learning Representations
  (ICLR)}, 2019.

\end{thebibliography}
\renewcommand{\bibfont}{\footnotesize}
\setlength{\bibsep}{1.5pt plus 0.2ex}

\end{document}